\RequirePackage{latexml}
\iflatexml
  \documentclass[10pt,letterpaper]{article}
  \usepackage{hyperref}
\usepackage{booktabs}
\usepackage{xcolor}
\usepackage{microtype}
\usepackage{authblk}
\newcommand{\AIRdate}[1]{\date{#1}}
\newcommand{\AIRhtmlvenue}{}
\newcommand{\AIRvenue}[1]{\renewcommand{\AIRhtmlvenue}{#1}}

\newlength{\AIRabstractteaserskip}
\newlength{\AIRafterteaserskip}
\newcommand{\AIRhtmlfrontmatter}{%
  \begingroup
  \def\PaperNote##1##2{\noindent\textsuperscript{\ensuremath{##1}}##2\par}%
  \PaperNotes
  \AIRrenderlinks{\par\noindent}\par
  \noindent\AIRhtmlvenue\par
  \endgroup}

\else
  \documentclass[10pt,letterpaper,logo]{thuair_xiangyu}
\fi
\newcommand{\arxivversion}{}

\usepackage[numbers,sort&compress]{natbib}

\makeatletter
\@ifundefined{iflatexml}{\newif\iflatexml}{}
\makeatother

\usepackage{xspace}
\usepackage{graphicx}
\usepackage{subcaption}
\usepackage{multirow}
\usepackage{makecell}
\usepackage[ruled,linesnumbered,noend]{algorithm2e}
\usepackage{amsmath}
\usepackage{amssymb}
\usepackage{enumitem}
\usepackage[capitalize]{cleveref}

\newcommand{\eg}{\emph{e.g.}\xspace}

\newcommand{\ie}{\emph{i.e.}\xspace}

\newcommand{\nickname}{OxyGen\xspace}

\input{tex/air_frontmatter}

\newcommand{\PaperNotes}{%
  \PaperNote{*}{Work done during internships at AIR, Tsinghua University.}%
  \PaperNote{\dagger}{Corresponding author: Ting Cao (tingcao@mail.tsinghua.edu.cn).}%
}

\iflatexml
  \newcommand{\AIRoperation}[3]{\item[$#1\to #2$] #3.}
  \newcommand{\AIRinterfaces}[1]{%
    \begin{description}[style=nextline,leftmargin=1.5em]#1\end{description}}
\else
  \makeatletter
  \newcommand{\AIRinterfaces}[1]{%
    \begingroup
    \gdef\AIRrows{}%
    \def\AIRoperation##1##2##3{%
      \ifx\AIRrows\@empty\else\g@addto@macro\AIRrows{\\}\fi
      \g@addto@macro\AIRrows{##1 &\to ##2 && \text{##3}}}%
    #1%
    \endgroup
    \begin{align*}\AIRrows\end{align*}}
  \makeatother
\fi

\newcommand{\AIRmaketitle}[1]{%
  \iflatexml
    \maketitle\AIRhtmlfrontmatter#1%
  \else
    #1\maketitle
  \fi}
\iflatexml
  \newenvironment{AIRteaser}{%
    \begingroup\captionsetup{font=small,skip=7pt}%
    \begin{figure}\centering\let\AIRteasercaption\caption
  }{\end{figure}\endgroup}
\else
  \newenvironment{AIRteaser}{%
    \begingroup\microtypesetup{expansion=false}%
    \captionsetup{font=small,skip=7pt,hypcap=false}%
    \par\vspace{\AIRabstractteaserskip}\noindent
    \begin{minipage}{\linewidth}\centering
    \newcommand{\AIRteasercaption}{\captionof{figure}}%
  }{\end{minipage}\par\endgroup\vspace{\AIRafterteaserskip}}
\fi

\title{\nickname: Unified KV Cache Management for VLA Inference under Multi-Task Parallelism}

\AIRdate{2026-10-13}
\AIRvenue{Accepted at the 40th Conference on Neural Information Processing Systems (NeurIPS 2026).}
\AIRsetupauthors
\hypersetup{
  pdftitle={OxyGen: Unified KV Cache Management for VLA Inference under Multi-Task Parallelism},
  pdfauthor={\AIRpdfauthors},
  pdfsubject={Unified KV cache management for VLA inference},
  pdfkeywords={VLA, KV cache, multi-task inference, OxyGen}
}

\makeatletter
\def\@fnsymbol#1{\ensuremath{\ifcase#1\or *\or \dagger\or \ddagger\or
    \mathsection\or \mathparagraph\or \|\or **\or \dagger\dagger
\or \ddagger\ddagger \else\@ctrerr\fi}}
\makeatother

\begin{document}

\AIRmaketitle{\begin{abstract}

  Embodied AI agents increasingly require parallel execution of multiple tasks, such as manipulation, conversation, and memory construction, from shared observations under distinct time constraints. Recent Mixture-of-Transformers (MoT) Vision-Language-Action Models (VLAs) support such heterogeneous outputs, yet existing inference systems incur redundant computation and resource contention when serving concurrent tasks on-device. We identify isolated KV cache management as the root cause.
  To address this, we propose \emph{unified KV cache management}, an inference design that treats the KV cache as a first-class shared resource across tasks and control cycles. This abstraction enables two key optimizations: \emph{cross-task KV sharing} eliminates redundant prefill of shared observations, while \emph{cross-frame continuous batching} batches unfinished language requests from different frames and advances them within a per-frame decoding budget.
  We implement this design in \nickname and evaluate $\pi_{0.5}$ on NVIDIA GeForce RTX 4090 and Jetson AGX Thor. \nickname achieves up to 3.7$\times$ speedup over isolated execution, delivering over 200~tokens/s language throughput and 70~Hz action frequency simultaneously on RTX 4090. We also validate the design on Xiaomi-Robotics-0 and StarVLA across different backbones and expert configurations. We further demonstrate the inference gains on a humanoid robot with on-board Jetson AGX Thor.

\end{abstract}
}

\begin{AIRteaser}
  \includegraphics[trim = 0 0 0 170, clip, width=\linewidth]{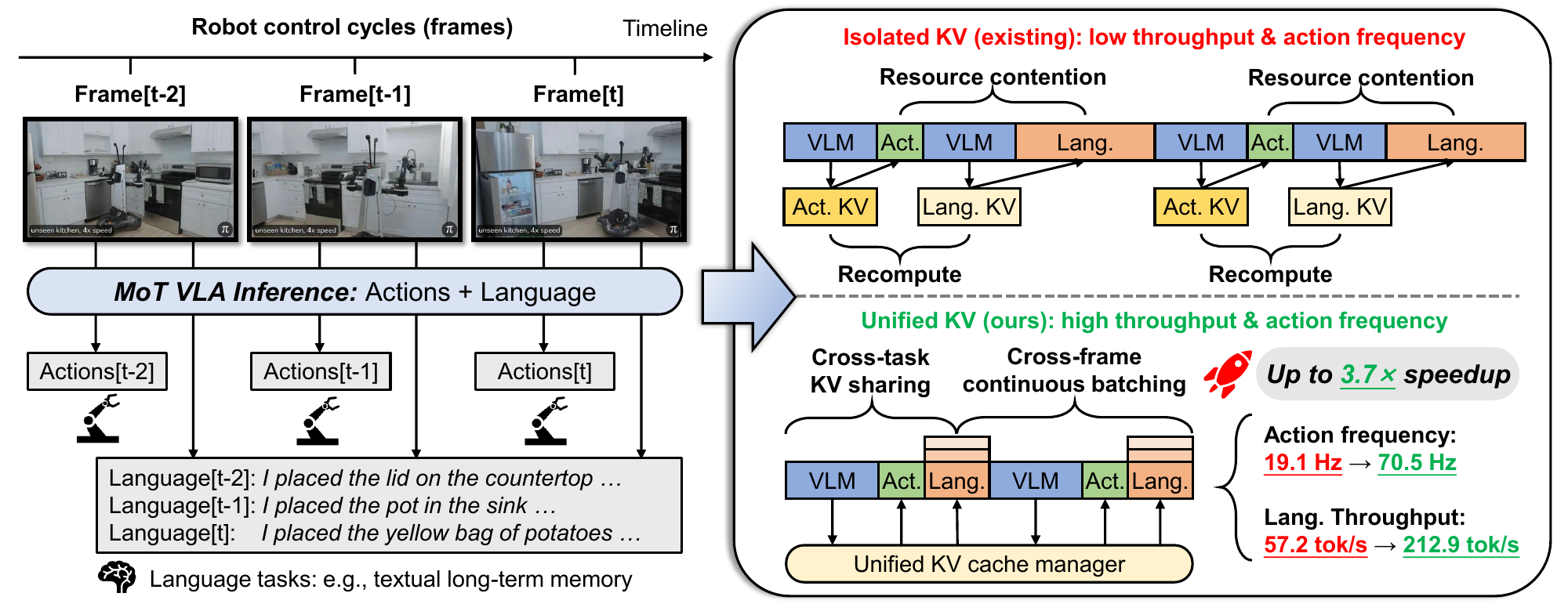}
  \AIRteasercaption{
    Left: An example of deploying a Mixture-of-Transformers (MoT) Vision-Language-Action (VLA) model for parallel multi-task inference: from per-frame observations, the VLA generates robot actions within each frame while continuously generating language-based memories across frames~\cite{torne2025mem}.
    Right: Comparison between paradigms of MoT VLA inference: existing systems manage the KV cache in isolation, slowing down inference through redundant computation and resource contention; our method adopts \emph{unified KV cache management}, achieving up to 3.7$\times$ speedup via cross-task KV sharing and cross-frame continuous batching.
  }
  \label{fig:teaser}
\end{AIRteaser}

\section{Introduction}
\label{sec:intro}

\ifdefined\arxivversion
\else
\begin{figure}[t]
  \centering
  \includegraphics[trim = 0 0 0 170, clip, width=\linewidth]{figures/design/teaser.pdf}
  \caption{
    Left: An example of deploying a Mixture-of-Transformers (MoT) Vision-Language-Action (VLA) model for parallel multi-task inference: from per-frame observations, the VLA generates robot actions within each frame while continuously generating language-based memories across frames~\cite{torne2025mem}.
    Right: Comparison between paradigms of MoT VLA inference: existing systems manage the KV cache in isolation, slowing down inference through redundant computation and resource contention; our method adopts \emph{unified KV cache management}, achieving up to 3.7$\times$ speedup via cross-task KV sharing and cross-frame continuous batching.
  }
  \label{fig:teaser}
\end{figure}
\fi

A long-standing aspiration in embodied AI is to develop agents that, much like humans, coordinate multiple tasks in parallel: conversing while manipulating objects \cite{figure_figure03_2025, 1x_neo_robot, lee2026modern_recipe, shi2025hi_robot}, or memorizing surroundings while navigating \cite{anwar2025remembr, rajvanshi2024saynav, gu2024conceptgraphs, kim2023topological}.
Consider an autonomous home robot such as the one in \cref{fig:teaser} \cite{1x_neo_robot, figure_figure03_2025,torne2025mem}: while manipulating, it must concurrently memorize environmental changes, narrate progress to the user, and plan ahead to update its schedule.
We refer to this setting as \textit{multi-task parallelism}: concurrent execution of temporally independent tasks from shared input, each under its own time constraints.

Recent advances in Vision-Language-Action Models (VLAs)~\cite{zitkovich2023rt, kim2024openvla, black2024pi_0, intelligence2025pi_05, bjorck2025gr00t, zhai2025wall_oss, li2024cogact, jiang2025galaxea, wen2025dexvla, bu2025univla, robotics2026xiaomi, cen2025rynnvla} and Mixture-of-Transformers (MoT) architectures~\cite{liang2024mixture,deng2025bagel,bi2025motus,nvidia2026cosmos3} have made strides toward this goal.
Conventional VLAs~\cite{zitkovich2023rt, kim2024openvla, bu2025univla, pertsch2025fast} focus on the action modality, so supporting concurrent tasks across modalities typically requires multi-model inference (\eg, running a VLA and a VLM concurrently), making on-device deployment difficult under limited hardware resources.
In contrast, recent MoT VLAs~\cite{intelligence2025pi_05,zhai2025wall_oss,robotics2026xiaomi} route different outputs to modality-specific experts, enabling a single model to perform both language-based tasks (\eg, planning) and action-based tasks (\eg, manipulation).
Yet this architectural multitasking capability does not automatically translate into efficient multi-task inference.

We find that existing systems~\cite{openpi2025, wallx2025, galaxeavla2025, xiaomirobotics2026} follow an inference paradigm similar to naive multi-model inference, which we term \emph{isolated execution}: each task executes a separate forward pass of the same model, even when tasks share the same input observations (\cref{fig:teaser}).
This leads to two inefficiencies.
(\romannumeral 1) Redundant computation: the shared observation is encoded repeatedly, producing identical KV cache entries for each task (1.4$\times$ slowdown in \cref{sec:ablation}).
(\romannumeral 2) Resource contention: even with a shared prefix, tasks compete for limited hardware resources and can delay one another despite their different time constraints. For example, action generation must keep pace with the robot's control loop, while language generation can span multiple control cycles. With KV sharing already enabled, cross-frame continuous batching provides a further 2.7$\times$ speedup in \cref{sec:ablation}.
Underlying both issues, we identify a common root cause: existing systems treat each task's KV cache in isolation, missing opportunities for sharing and coordinated scheduling.

Although KV cache reuse~\cite{kwon2023efficient,zheng2024sglang,yao2025cacheblend,yang2025kvshare,mi2026kvfetcher} and continuous batching~\cite{kwon2023efficient,zheng2024sglang} are well studied in LLM serving, directly applying them to MoT VLA inference leaves two challenges.
(\romannumeral 1)~\emph{Heterogeneous expert KV semantics}: LLM prefix caching reuses context within an autoregressive generation path. In MoT VLAs, however, one prefill result must support separately weighted experts: the action expert repeatedly reads the prefix during denoising, while the language expert maintains and extends its own cache across frames. The system must coordinate prefill reuse across these different generation processes and cache lifetimes.
(\romannumeral 2)~\emph{Frame-bounded language decoding}: standard continuous batching schedules autoregressive requests without coordinating their progress with recurring action inference. The robot's control loop instead requires language decoding to advance within each frame's time budget, retain its state, and resume across frames without recomputation.

To address this, we propose \textit{unified KV cache management}, which exposes the \textit{KV cache as a first-class, unified resource managed across tasks and over time.}
We realize it in \nickname, an efficient multi-task inference system for MoT VLAs on robotic platforms, with two optimizations: (\romannumeral 1) \emph{cross-task KV sharing} encodes the shared observation once and reuses the resulting KV cache across experts, eliminating redundant prefill; (\romannumeral 2) \emph{cross-frame continuous batching} groups in-flight language requests from different frames into a single decoding batch, advancing each within a per-frame decoding budget.
Together, these optimizations reduce redundant computation and amortize language decoding across requests, improving both action frequency and aggregate language throughput.

We implement \nickname for $\pi_{0.5}$~\cite{intelligence2025pi_05} atop openpi~\cite{openpi2025}, its official inference framework, with both JAX and PyTorch backends.
Across the LIBERO, DROID, and ALOHA configurations on NVIDIA GeForce RTX 4090 and Jetson AGX Thor, two representative platforms for on-device VLA inference~\cite{jiang2026fast,ma2025running,black2024pi_0}, \nickname achieves up to $3.7\times$ speedup in action frequency and language throughput, delivering over 200~tokens/s and 70~Hz simultaneously on RTX 4090.
We further validate \nickname on a humanoid robot with on-board Jetson AGX Thor, where it reduces inference latency from 1030 to 393~ms. Once actions are dispatched, language decoding can run during physical execution, hiding its latency (\cref{sec:exp-realrobot}).
We also evaluate \nickname on Xiaomi-Robotics-0 and StarVLA with up to four experts, demonstrating its applicability across backbones and expert configurations.

In summary, our contributions are threefold:
\begin{itemize}[leftmargin=1.2em,topsep=2pt,itemsep=1pt]
  \item We formulate multi-task parallelism as a target inference scenario for MoT VLAs, and identify isolated KV cache as the root cause of inefficiency in existing systems.
  \item We propose unified KV cache management, an inference design that treats KV cache as a shared resource across tasks and over time, enabling cross-task KV sharing and cross-frame continuous batching.
  \item We implement this design for $\pi_{0.5}$ and evaluate it on common robotic GPUs (GeForce RTX 4090 and Jetson AGX Thor), demonstrating up to $3.7\times$ speedup in action frequency and language throughput. We further validate the design across additional MoT VLA backbones and multi-expert configurations.
\end{itemize}

\section{Related Work}

\paragraph{MoT VLAs.}
\label{sec:vla-arch}
Earlier VLA paradigms focus on action generation: \emph{discrete} VLAs~\cite{zitkovich2023rt, kim2024openvla} generate actions autoregressively as language tokens, while \emph{continuous} VLAs~\cite{li2024cogact, black2024pi_0, bjorck2025gr00t} attach a lightweight diffusion or flow-matching action module for high-frequency control.
We instead target recent MoT VLAs (\eg, $\pi_{0.5}$~\cite{intelligence2025pi_05}, WALL-OSS~\cite{zhai2025wall_oss}, Xiaomi-Robotics-0~\cite{robotics2026xiaomi}), which route different output modalities to separate experts while sharing a common Vision-Language Model (VLM) backbone~\cite{liang2024mixture}, enabling a single model to support concurrent action- and language-based tasks.
Despite this architectural multitasking capability, existing MoT VLA inference systems still execute tasks through independent forward passes, repeating shared computation and leaving tasks with different timing requirements to compete for resources.
\nickname addresses this gap by managing KV caches across tasks and control cycles, enabling both computation reuse and coordinated execution.

\paragraph{VLA inference systems and optimizations.}
\label{sec:vla-optimization}
Recent inference frameworks~\cite{xu2026embodiedcpp,phyai2026,nguyen2026vlacpp} provide unified runtimes for deploying different VLA architectures across heterogeneous hardware. Specific optimization techniques target two levels.
(\romannumeral 1)~\emph{Model-level}: VLAs inherit many VLM optimizations, including compression~\cite{wang2025bitvla, fang2025sqapvla, yang2025efficientvla, park2024qail}, token pruning~\cite{tan2025flashvla, li2025spvla, yang2025efficientvla, jiang2025lightvla}, layer skipping~\cite{zhang2025molevla, yue2024deervla, reuss2025flower, shukor2025smolvla}, action token reuse~\cite{tan2025flashvla, xu2025vlacache}, KV cache pruning~\cite{xu2025kvefficientvla}, and graph optimization~\cite{ma2025running}. KV-Efficient VLA~\cite{xu2025kvefficientvla} selectively reduces KV computation within model inference, whereas \nickname manages KV reuse and execution across concurrent tasks.
(\romannumeral 2)~\emph{Execution-level}: asynchronous execution methods~\cite{shukor2025smolvla,black2025realtimechunk,zhao2025vlarail,black2025trainingrtc,tang2025vlash} overlap action inference with physical execution while maintaining continuity across action chunks.
\nickname is largely orthogonal to both directions: it targets computation reuse and execution coordination across concurrent inference tasks. It could be combined with model-level optimizations to reduce inference cost and with asynchronous execution methods to overlap multi-task inference with robot execution.

\paragraph{KV cache management in LLM serving.}
KV cache management is widely studied in LLM serving: prefix caching in vLLM~\cite{kwon2023efficient} and SGLang~\cite{zheng2024sglang} avoids recomputation when requests share prefix tokens, with extensions to non-prefix~\cite{yao2025cacheblend,yang2025kvshare} and cross-model~\cite{fu2025cache,liu2024droidspeak} reuse.
Continuous batching further improves throughput by batching active requests as others arrive or complete.
Applying these techniques to MoT VLA inference requires reusing a prefill result across a flow-matching expert that reads the prefix during denoising and an autoregressive expert that maintains its own evolving cache, while scheduling language decoding within the time budget of recurring action inference.
Simply enabling prefix caching and continuous batching in an LLM server does not provide this coordination.
\nickname addresses these requirements through unified KV cache management across experts and control cycles.

\section{Method}
\label{sec:method}

\begin{figure}[t]
  \centering
  \includegraphics[trim = 0 0 0 135, clip, width=\linewidth]{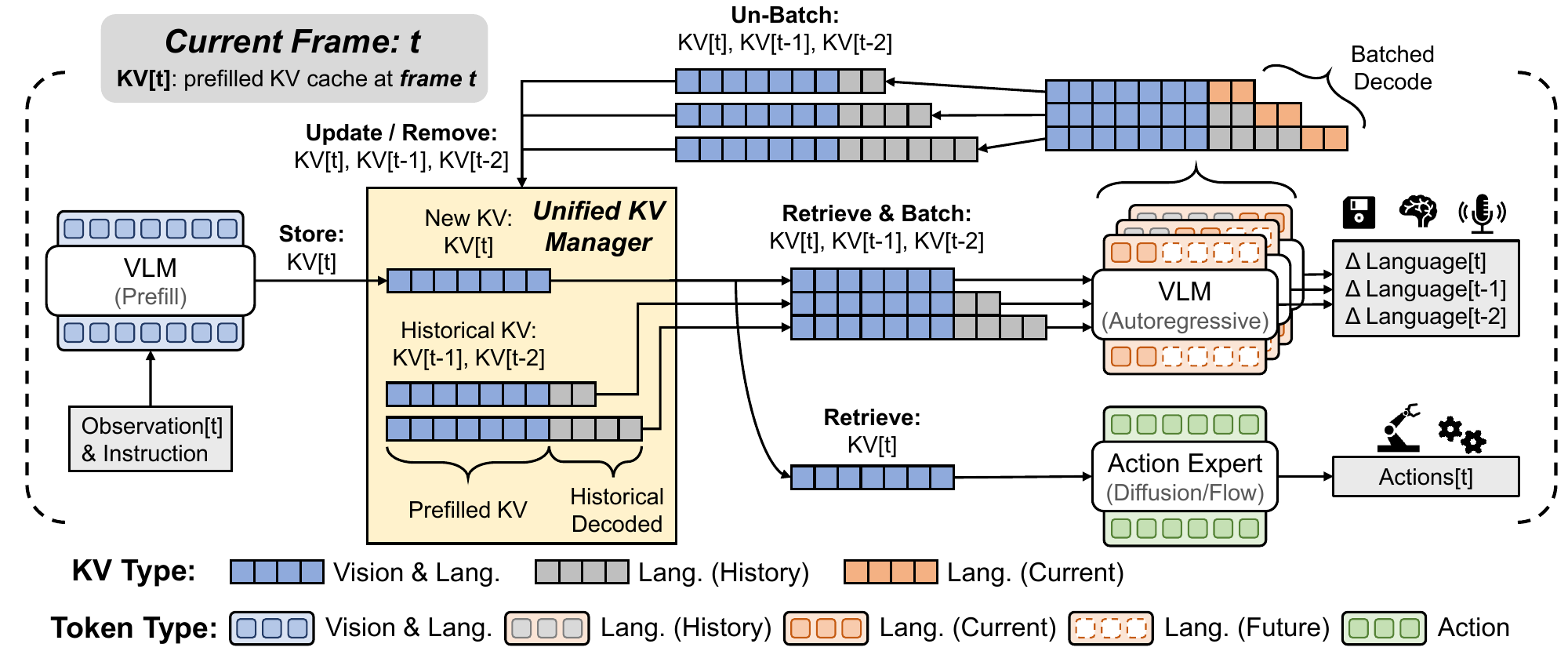}
  \caption{KV-centric dataflow at frame $t$ with a unified KV cache manager. KV[t] tracks the cache originating from frame $t$: its prefill portion is defined in \cref{eq:kv-cache}, and language decoding adds token KVs. $\Delta$Language[t] represents incremental language tokens in $\mathbf{y}_t$.}
  \label{fig:dataflow}
\end{figure}

We propose \nickname, an inference system for MoT VLAs that achieves efficient multi-task parallelism through \emph{unified KV cache management}.
The key insight is that the KV cache, produced by the shared VLM backbone from a common observation, is a natural locus for both computation reuse and execution coordination.

\subsection{Preliminaries and Problem Formulation}
\label{sec:formulation}

\paragraph{MoT VLA inference.}
We consider a generic multi-task embodied agent based on a MoT VLA, such as $\pi_{0.5}$~\cite{intelligence2025pi_05}.
At frame (\ie, control cycle) $t$, the agent observes $\mathbf{o}_t$, which contains visual inputs for this frame and a language instruction.
A MoT VLA factorizes inference into prefill with the modality-agnostic backbone and generation with modality-specific experts.
The prefill phase is formulated as:
\begin{equation}
  \bigl\{(\mathbf{h}_{t,l}, \mathbf{K}_{t,l}, \mathbf{V}_{t,l})\bigr\}_{l=1}^{L} = \Theta_{\text{VLM}}(\mathbf{o}_t), \quad \mathcal{K}_t = \bigl\{(\mathbf{K}_{t,l},\, \mathbf{V}_{t,l})\bigr\}_{l=1}^{L},
  \label{eq:kv-cache}
\end{equation}
where $\Theta_{\text{VLM}}$ denotes the parameters of the VLM backbone, $L$ is the number of transformer layers in the VLM, $\mathbf{h}_{t,l}, \mathbf{K}_{t,l}, \mathbf{V}_{t,l}$ are hidden states, keys, and values of VLM layer $l$, and $\mathcal{K}_t$ is the KV cache produced from $\mathbf{o}_t$.
Crucially, $\mathcal{K}_t$ is modality-agnostic: it encodes the observation and can be consumed by multiple experts, without committing to a specific output modality.

Given the shared $\mathcal{K}_t$, the MoT VLA runs multiple experts independently. In this paper, we focus on two representative experts: an \emph{action expert} that generates an action chunk $\mathbf{A}_t = \{\mathbf{a}_{t,i}\}_{i=1}^{H}$ (\ie, low-level control commands for a horizon of $H$), and a \emph{language expert} that generates text tokens $\mathbf{y}_t = \{y_{t,j}\}_{j=1}^{N}$ (\eg, memory or QA with a maximum token budget of $N$). Since $\mathcal{K}_t$ encapsulates the visual-language information from $\mathbf{o}_t$, both experts can generate their outputs conditioned on $\mathcal{K}_t$ instead of directly on $\mathbf{o}_t$:
\begin{equation}
  p_{\Theta_{\text{Act}}}(\mathbf{A}_t \mid \mathcal{K}_t), \quad
  p_{\Theta_{\text{Lang}}}(\mathbf{y}_t \mid \mathcal{K}_t),
  \label{eq:distribution}
\end{equation}
where $\Theta_{\text{Act}}$ and $\Theta_{\text{Lang}}$ (usually the language backbone in the VLM) parameterize the action and language distributions, respectively. Concretely, the language expert decodes $\mathbf{y}_t$ autoregressively conditioned on $\mathcal{K}_t$, while the action expert generates the chunk $\mathbf{A}_t$ jointly via $S$ denoising steps over $\mathcal{K}_t$ (implemented as diffusion or flow matching); full formulations are deferred to \cref{sec:appendix-generation}.

\paragraph{Multi-task parallelism with asymmetric deadlines.}
At each frame $t$, the agent serves concurrent action and language tasks with asymmetric deadlines. Action generation must meet a per-frame deadline to sustain a target action frequency $f_{\min}$, while language generation can span multiple frames under a soft deadline. Our objective is to maximize language throughput while meeting the action deadline.
Let $f$ and $\tau$ denote action frequency and language throughput at steady state. With one new request per frame, each frame produces $H$ actions and advances a batch of $B$ language requests by $k$ decoding steps. With frame latency $T$, we have $f=H/T$ and $\tau=Bk/T$, giving:
\begin{equation}
  \max \quad \tau = \frac{Bk}{T} \quad \text{s.t.} \quad f = \frac{H}{T} \geq f_{\min},
  \label{eq:objective}
\end{equation}
Here, action frequency counts generated actions; the replanning rate is $1/T$. The objective suggests two optimization directions: reducing frame latency and increasing the number of tokens decoded per frame. Completing language generation within each frame delays the next action update, while limiting decoding to a few tokens without batching sacrifices language throughput. \nickname addresses both through unified KV cache management across tasks and frames.

\subsection{Unified KV Cache Manager}
\label{sec:unified-kv}

\nickname abstracts the KV cache as a shared resource across tasks and frames, managed by a \emph{unified KV cache manager} $\mathcal{M}$.
The manager supports two key optimizations: \textbf{cross-task KV sharing} reuses one prefill result $\mathcal{K}_t$ across action and language experts within frame $t$, eliminating redundant prefill; \textbf{cross-frame continuous batching} groups in-flight language requests from different frames into a single decoding batch, allowing language generation to progress across successive action updates.
To support both, $\mathcal{M}$ maintains a generation state for each in-flight request, tracking its KV cache and decoded tokens.

\paragraph{Resumable generation state.}
Cross-task KV sharing lets both experts reuse a single prefill computation.
To resume language generation across frames, we maintain a separate \emph{incremental state} for each language request:
\begin{equation}
  \sigma_t = \bigl(\mathcal{K}_t,\; \mathbf{y}_t,\; \delta_t\bigr),
  \label{eq:state}
\end{equation}
where $\mathbf{y}_t$ stores the generated tokens and $\delta_t \in \{0,1\}$ indicates termination when EOS is emitted or the maximum length $N$ is reached.
Within $\sigma_t$, $\mathcal{K}_t$ denotes the language request's cache, containing the prefill from frame $t$ and the token KVs added during decoding. The action expert uses the prefill cache independently for denoising.
Crucially, $\sigma_t$ preserves the generation context needed to resume decoding without recomputation. Cache allocation and reuse are described in \cref{sec:appendix-memory}.
The manager $\mathcal{M}$ exposes standard CRUD operations (\textsc{Store}, \textsc{Retrieve}, \textsc{Update}, \textsc{Remove}) on per-request states $\sigma_t$, and maintains the set of active requests $\mathcal{R} = \{r_{t_1}, \ldots, r_{t_m}\}$ whose generation has not yet terminated.
For batched decoding, $\mathcal{M}$ stacks the active states along the batch dimension into a batched state $\hat{\sigma}$, so that the VLM advances all $m = |\mathcal{R}|$ requests in a single forward pass; the formal interface and definition of $\hat{\sigma}$ are given in \cref{sec:appendix-manager}.

\begin{figure}[t]
  \centering
  \includegraphics[trim = 0 0 0 105, clip, width=\linewidth]{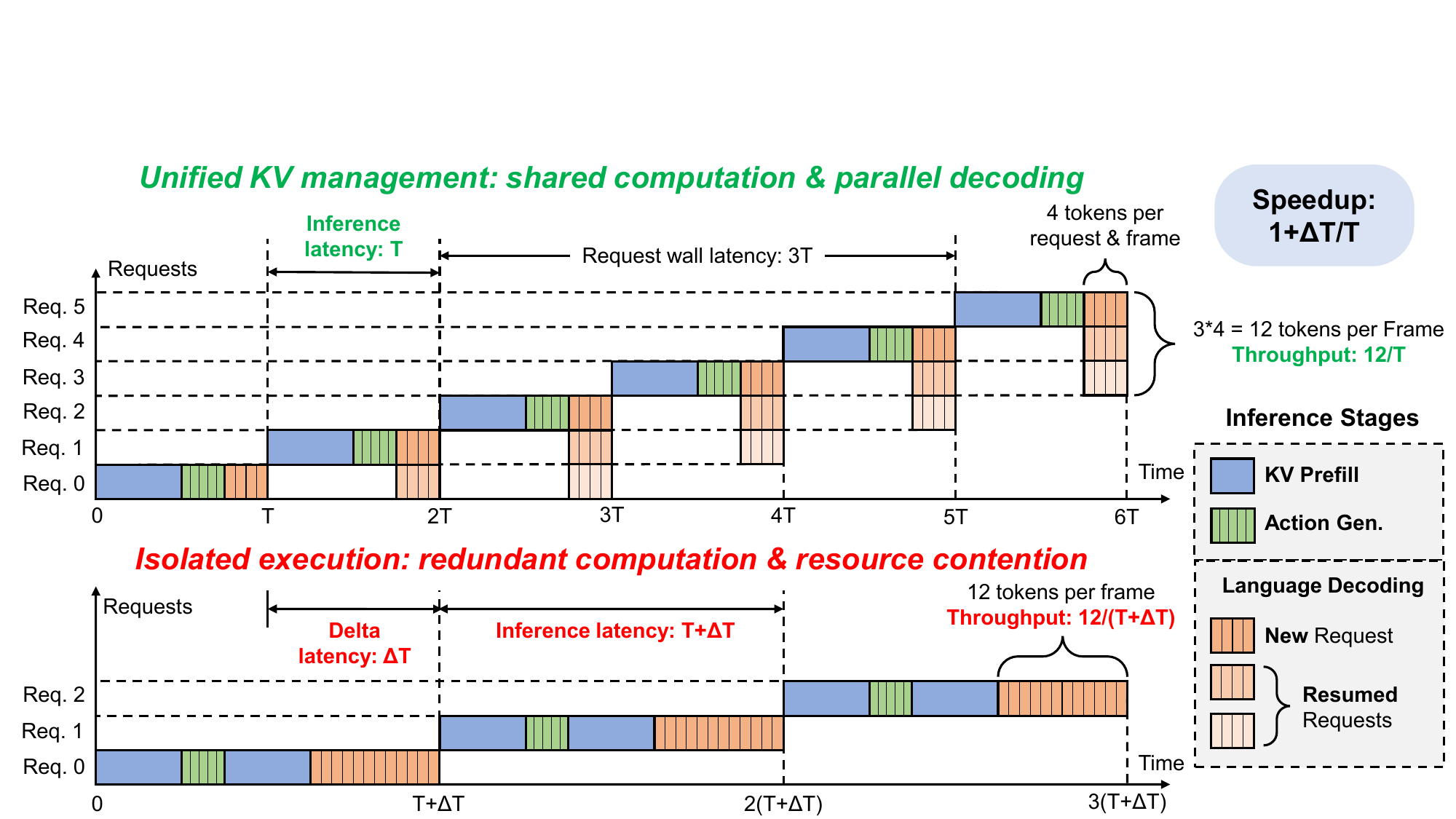}
  \caption{Timeline comparison of \nickname and isolated execution with $N=12$ tokens per request. After warmup, \nickname advances $B=3$ active requests by $k=4$ tokens each per frame. Both systems produce 12 language tokens per frame, but \nickname reduces frame latency from $T+\Delta T$ to $T$, increasing both action frequency and aggregate language throughput by a factor of $1+\Delta T/T$. Each language request in \nickname spans three frames, completing in $3T$.}
  \label{fig:timeline}
\end{figure}

\paragraph{Per-frame execution flow.}
The two optimizations compose into a two-stage per-frame execution.
First, the system runs prefill once on the new observation $\mathbf{o}_t$ to obtain $\mathcal{K}_t$ (\cref{eq:kv-cache}); the manager fans this cache out to the action expert (which denoises $\mathbf{A}_t$ over $S$ steps) and to a freshly initialized language state $\sigma_t$ stored under request ID $r_t$, eliminating the redundant prefill required by isolated execution.
Second, $r_t$ joins the active set $\mathcal{R}$, and $\mathcal{M}$ retrieves and batches all $m$ active states into $\hat{\sigma}$ for $k$ steps of batched autoregressive decoding; finished requests are then evicted and the rest are persisted with their incremental tokens $\mathbf{y}_{t_i}$ for the next frame.
\cref{alg:per-frame} in \cref{sec:appendix-algorithm} formalizes this flow.
\cref{fig:timeline} illustrates the combined effect: cross-task KV sharing eliminates redundant prefill, while cross-frame continuous batching amortizes language decoding across requests from different frames.

The decoding budget $k$ determines how far each language request advances per frame. With one new request per frame and a fixed request length, a smaller $k$ spreads generation over more frames and increases the active batch size. This can improve both action frequency and aggregate language throughput, at the cost of longer completion times for individual language requests. For simplicity, our current implementation uses a fixed $k$. In deployment, $k$ could be chosen to meet action-frequency and language-response requirements by profiling the expected workload on the target hardware.

\paragraph{Discussion: generalization to richer settings.}
We instantiate \nickname with two experts and standard autoregressive language decoding in $\pi_{0.5}$. The manager interface also supports additional experts and offers a basis for other decoding schemes.
\textbf{(i) More experts.}
The manager can fan out the shared prefill cache $\mathcal{K}_t$ to additional experts that use the same backbone's KV representation, while maintaining a separate generation state for each autoregressive request. We evaluate this execution flow with two, three, and four experts in \cref{sec:exp-models}. For language tasks, our textual-memory example applies a LoRA adapter~\cite{hu2022lora} only to the language continuation, preserving the shared prefill (\cref{sec:appendix-memory-demo}). Cross-model KV transfer~\cite{liu2024droidspeak,fu2025cache} and multi-LoRA serving~\cite{sheng2024slora,chen2024punica} offer potential extensions to tasks using different language models or distinct adapters on a shared backbone, respectively.
\textbf{(ii) Parallel decoding.}
Although our implementation uses standard autoregressive decoding, the manager's resumable-state interface does not require one token per forward pass. It could also support speculative or parallel decoding~\cite{cai2024medusa,li2024eagle,zhang2025hass}, with decoding-specific state retained as needed between frames.

\section{Experiments}
\label{sec:experiments}

\subsection{Experimental Setup}
\label{sec:setup}

\subsubsection{Models, Configurations, and Hardware}
We evaluate \nickname with $\pi_{0.5}$~\cite{intelligence2025pi_05} on NVIDIA GeForce RTX 4090 and Jetson AGX Thor, two representative platforms for on-device VLA inference~\cite{jiang2026fast,ma2025running,black2024pi_0}.
We measure inference speed under three configurations derived from LIBERO~\cite{liu2023libero}, DROID~\cite{khazatsky2024droid}, and ALOHA~\cite{zhao2023aloha}, matching their camera views, image resolution, and control dimensionality.
We evaluate task success on LIBERO using the officially released $\pi_{0.5}$-LIBERO checkpoint.
For these inference-speed experiments, we focus on LIBERO in the main text and report the full results in \cref{sec:appendix-e2e}.

To assess generality across backbones and expert configurations, we additionally evaluate Xiaomi-Robotics-0~\cite{robotics2026xiaomi} and models built with StarVLA~\cite{starvla2026} on RTX 4090 (\cref{sec:exp-models}).

\subsubsection{Baselines}
We compare \nickname against openpi~\cite{openpi2025}, the official inference framework for $\pi_{0.5}$, running in two execution modes:

\paragraph{Sequential isolated execution} (the main baseline, denoted as ``Baseline'') follows the standard inference paradigm of existing systems~\cite{openpi2025, wallx2025, galaxeavla2025, xiaomirobotics2026}: each task (action and language generation) runs independently and sequentially within each frame.
Since openpi does not release code for the language generation described in the $\pi_{0.5}$ paper, we implement this baseline following a community reproduction\footnote{\url{https://github.com/BrunoFANG1/openpi_subtask_generation}.}.

\paragraph{Parallel isolated execution} (an additional baseline, denoted as ``Parallel'') runs action and language generation in separate processes on the same GPU.
We implement this for openpi via CUDA Multi-Process Service (MPS).

\subsubsection{Metrics}
We measure the following metrics to capture both inference speed and deployment efficiency.
\textbf{Action frequency} (Hz): action horizon divided by mean end-to-end frame latency, $H/T$. For $\pi_{0.5}$, we set $H=10$ and $S=10$, unless otherwise specified.
\textbf{Language throughput} (tokens/s): total language tokens generated per second across active requests.
\textbf{Average batch size}: average number of concurrent requests during language decoding, reflecting the parallelism of language generation.
\textbf{Memory} (GiB), \textbf{Power} (W), and \textbf{Energy/Request} (J): metrics reflecting the system's memory and energy efficiency for on-device deployment.

\subsection{End-to-End Results}
\label{sec:exp-e2e}

\begin{figure}[t]
  \centering
  \includegraphics[width=\linewidth]{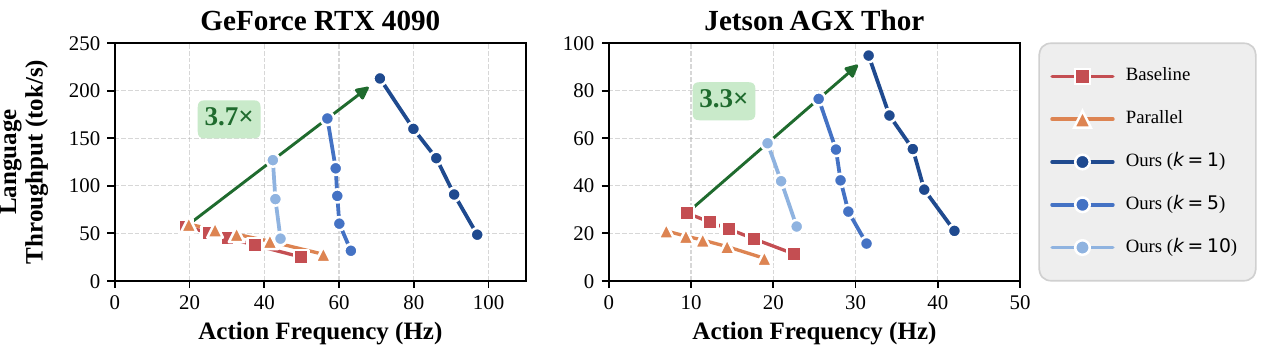}
  \caption{Action-frequency vs.\ language-throughput tradeoff on the LIBERO configuration. We sweep decoding steps per frame $k \in \{1, 5, 10\}$ and maximum total decoding steps $N \in \{5, 10, 15, 20, 30\}$; each \nickname line fixes $k$ and varies $N$ (markers are individual settings). Each baseline curve must trade one axis for the other as $N$ grows; \nickname improves both metrics by up to $3.7\times$ over sequential isolated execution at matched output lengths.}
  \label{fig:e2e-tradeoff}
\end{figure}

\begin{figure}[t]
  \centering
  \includegraphics[width=\linewidth]{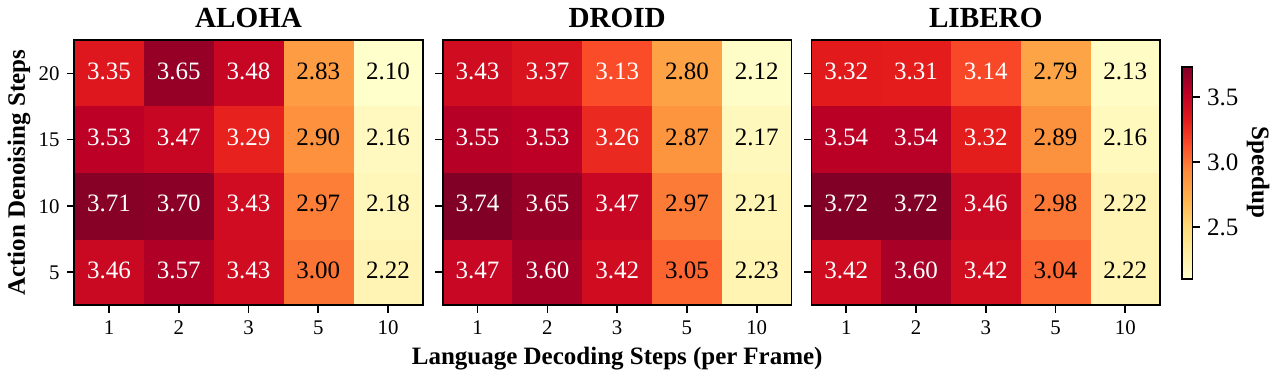}
  \caption{Speedup ratio for action frequency of \nickname vs.\ baseline across LIBERO, ALOHA, and DROID configurations (subplots), action denoising steps $S$ (y-axis), and decoding steps per frame $k$ (x-axis), with total decoding steps fixed at $N=30$. The speedup is primarily affected by $k$, while the configuration choice and $S$ have smaller impact.}
  \label{fig:e2e-speedup-heatmap}
\end{figure}

We compare \nickname against both sequential and parallel isolated execution (denoted as ``Baseline'' and ``Parallel'') across settings that vary the decoding steps per frame $k \in \{1, 5, 10\}$ and the maximum total decoding steps $N$. One new observation and request arrive each frame. \cref{fig:e2e-tradeoff} sweeps the joint action-frequency / language-throughput operating space on the LIBERO configuration; each \nickname curve fixes $k$ and varies $N$, while the baseline curves vary $N$. \cref{fig:e2e-speedup-heatmap} reports the action-frequency speedup ratio over the baseline as we vary $k$ at fixed $N=30$.

\paragraph{Key observations.}
(\romannumeral 1)~\nickname extends the action-frequency / language-throughput Pareto frontier. As language outputs grow longer, both isolated-execution baselines gain language throughput at the expense of action frequency. At matched output lengths, \nickname improves both metrics by 1.2--3.7$\times$ over sequential isolated execution, supporting faster action updates while completing the same language workload.
(\romannumeral 2)~\nickname achieves higher speedup with larger total decoding steps $N$ and smaller decoding steps per frame $k$. With one new request per frame, the tested settings yield a steady-state batch size of $B=N/k$. Larger batches amortize language decoding across more requests, reducing its per-frame cost compared with isolated execution.
(\romannumeral 3)~MPS parallelization offers limited gains on RTX 4090 and reduces performance on Jetson AGX Thor: the two processes still repeat prefix computation and compete for GPU resources. \nickname removes redundant prefill and batches language requests across frames, addressing both sources of inefficiency.

\subsection{Ablation Study}
\label{sec:ablation}

\begin{figure}[t]
  \centering
  \includegraphics[width=\linewidth]{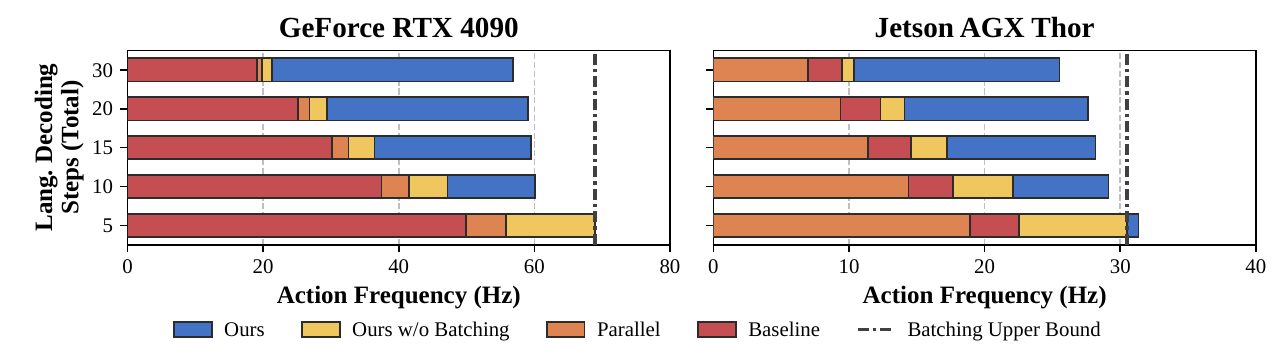}
  \caption{Ablation of action frequency on LIBERO across language decoding steps (5 steps per frame), measured on GeForce RTX 4090 (left) and Jetson AGX Thor (right). Cross-task KV sharing (``Ours w/o Batching'') provides the initial speedup for short decoding steps, while cross-frame continuous batching (``Ours'') maintains high action frequency as decoding steps increase, on both platforms. The dash-dotted line marks the single-frame decoding oracle upper bound.}
  \label{fig:ablation-speedup}
\end{figure}

We ablate the two core optimizations to understand their individual contributions.
Starting from sequential isolated execution (``Baseline''), we incrementally enable (A)~cross-task KV sharing (``Ours w/o Batching'') and (B)~cross-frame continuous batching (``Ours''), measuring action frequency. We also include multi-process parallelization (``Parallel'') for comparison.
For a more comprehensive view, we also measure a single-frame decoding oracle (``Batching Upper Bound'') that decodes only the per-frame budget of $k$ language tokens and drops the remaining $N-k$ tokens, so no language request carries over to later frames. This estimates the action-frequency ceiling when cross-frame language backlog is removed. \cref{fig:ablation-speedup} shows the action frequency on LIBERO across language decoding steps (5 steps per frame) on both GeForce RTX 4090 and Jetson AGX Thor.

\paragraph{Key observations.}
(\romannumeral 1)~Cross-task KV sharing eliminates redundant prefill of the same observation for different tasks. On RTX 4090 with LIBERO and $S=10$, it reduces frame latency from 200.3 to 145.1~ms at $N=5$, providing a 1.4$\times$ speedup before batching is enabled.
(\romannumeral 2)~Cross-frame continuous batching maintains nearly constant action frequency as language outputs grow longer, sustaining around 60~Hz on RTX 4090 and 27~Hz on Jetson AGX Thor. In contrast, execution without batching slows substantially as decoding length increases. With KV sharing already enabled, batching at $N=30,k=5$ reduces frame latency from 469.3 to 175.8~ms on RTX 4090, providing a further 2.7$\times$ speedup and increasing action frequency from 21.3 to 56.9~Hz.

The two optimizations address complementary costs. On RTX 4090, KV sharing saves approximately 55--58~ms per frame by removing duplicate prefill. As language outputs grow longer, decoding accounts for more of the frame latency, making continuous batching increasingly beneficial. \cref{sec:appendix-small-n} provides a stage-level breakdown for short language outputs.

\subsection{Evaluation across Backbones and Experts}
\label{sec:exp-models}

To evaluate \nickname beyond $\pi_{0.5}$, we test Xiaomi-Robotics-0~\cite{robotics2026xiaomi} and StarVLA's~\cite{starvla2026} Qwen3VL-PI\_v3 model on RTX 4090. The latter combines a Qwen3-VL-4B backbone~\cite{bai2025qwen3vl} with a PI\_v3 action head. We first evaluate one action expert alongside the language expert, then increase the number of action experts in StarVLA. We report frame latency and speedup over isolated execution within each configuration. With fixed output lengths and one new request per frame, this speedup also applies to action frequency and aggregate language throughput. \cref{sec:appendix-models} provides the experimental settings and additional results.

With one action expert and one language expert, \nickname accelerates Xiaomi-Robotics-0 and Qwen3VL-PI\_v3 by 3.23$\times$ and 3.60$\times$ at $N=30,k=5$, respectively (\cref{tab:model-transfer}). Reducing $k$ to 1 increases these gains to 5.83$\times$ and 6.64$\times$. Both backbones follow the trend observed with $\pi_{0.5}$: longer language outputs and smaller per-frame decoding budgets create larger batches, increasing the benefit of cross-frame continuous batching. \nickname reduces frame latency in every tested setting with $N>k$.

\begin{table}[t]
  \centering
  \small
  \setlength{\tabcolsep}{4pt}
  \renewcommand{\arraystretch}{1.05}
  \caption{Transfer across backbones and experts on RTX 4090, with $N=30$. $E$ counts all language and action experts. Speedup is relative to isolated execution within each configuration.}
  \label{tab:model-transfer}
  \begin{tabular}{lrrrr}
    \toprule
    Configuration & $E$ & Baseline (ms) & $k=1$ & $k=5$ \\
    \midrule
    Xiaomi-Robotics-0 & 2 & 562.8 & \textbf{5.83$\boldsymbol{\times}$} & 3.23$\times$ \\
    Qwen3VL-PI\_v3 (StarVLA) & 2 & 478.0 & \textbf{6.64$\boldsymbol{\times}$} & 3.60$\times$ \\
    \quad + GR00T & 3 & 515.8 & \textbf{6.05$\boldsymbol{\times}$} & 3.53$\times$ \\
    \quad + GR00T + OFT & 4 & 541.2 & \textbf{6.30$\boldsymbol{\times}$} & 3.69$\times$ \\
    \bottomrule
  \end{tabular}
\end{table}

We next add GR00T and OFT action experts to StarVLA, increasing the total expert count from two to four while reusing the same Qwen3-VL prefill. At $N=k=5$, where each language request finishes within one frame, \nickname's speedup over isolated execution increases from 1.23$\times$ to 1.40$\times$ and 1.59$\times$ with two, three, and four experts, respectively. Additional experts increase the amount of redundant prefill eliminated. With cross-frame batching at $N=30,k=5$, the three- and four-expert configurations achieve 3.53$\times$ and 3.69$\times$ speedups, respectively.

We further evaluate Qwen3.5~\cite{qwen352026} backbones from 0.8B to 9B with matching action heads through StarVLA. On RTX 4090, the $N=30,k=5$ speedup ranges from 3.58$\times$ to 3.79$\times$ across model sizes. \cref{tab:model-scaling} reports the full results.

\subsection{Performance under Different Workloads}
\label{sec:exp-workload}

\begin{figure}[t]
  \centering
  \includegraphics[width=\linewidth]{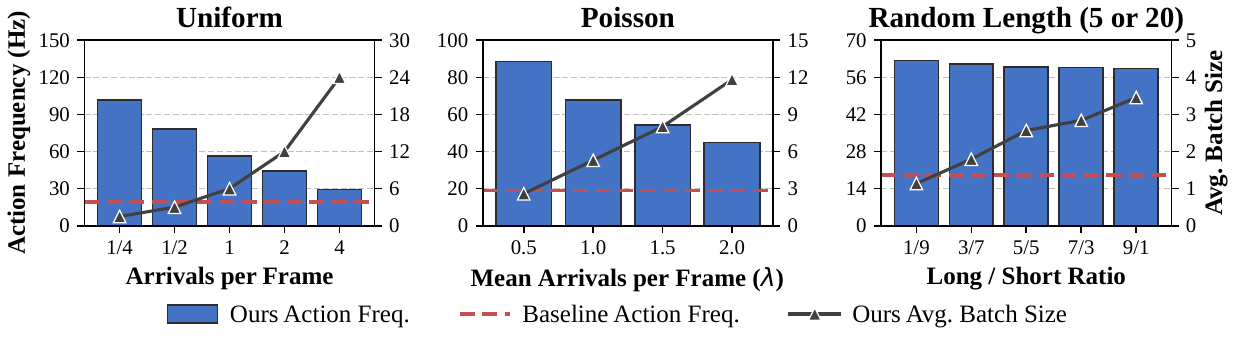}
  \caption{Action frequency and average language batch size under different request arrival patterns. Dashed lines show the isolated-execution baseline with one new request per frame and 30 language tokens per request.}
  \label{fig:workload-sweep}
\end{figure}

Using $\pi_{0.5}$-LIBERO on RTX 4090 with $k=5$, we evaluate uniform arrivals at different rates, Poisson arrivals with varying intensity $\lambda$, and mixtures of short and long language requests. \cref{fig:workload-sweep} reports action frequency and average language batch size across these workloads. The dashed line shows the isolated-execution baseline with one new request per frame and 30 language tokens per request.

\paragraph{Key observations.}
(\romannumeral 1)~Under uniform and Poisson arrivals, \nickname forms larger language batches as the arrival rate increases, improving language throughput. Action frequency decreases as more observations require processing within each frame.
(\romannumeral 2)~With mixed request lengths, \nickname maintains an action frequency of approximately 59--62~Hz while the average language batch size varies from 1.1 to 3.5. Requests of different lengths can therefore share the decoding batch with little change in action frequency.

\subsection{Action Quality and Resource Efficiency}
\label{sec:exp-quality-overhead}

We evaluate action quality using the official $\pi_{0.5}$-LIBERO checkpoint across four LIBERO suites. Over 10 seeds and 20,000 rollouts, \nickname achieves a mean success rate of 96.76\% (95\% CI: 96.55--96.97\%), compared with the 96.85\% reported by openpi. These results show comparable overall task success while accelerating inference. \cref{sec:appendix-quality} reports the per-suite results and evaluation protocol.

On RTX 4090, \nickname adds 15\% peak memory overhead over isolated execution while reducing average power by up to 47\% and energy per request by up to 78\%. Batched decoding amortizes VLM weight accesses across language requests. Naive parallelization nearly doubles peak memory and increases energy per request by 3\% (\cref{tab:overhead} in \cref{sec:appendix-overhead}).

Additional profiling and sustained-load tests examine KV-management costs and runtime stability in \cref{sec:appendix-manager-overhead,sec:appendix-stability}.

\subsection{Real-Robot Deployment}
\label{sec:exp-realrobot}

To evaluate on-board inference in a real deployment, we run \nickname with $\pi_{0.5}$ on a humanoid robot equipped with Jetson AGX Thor, generating actions and language from shared observations. We use three RGB views at $224{\times}224$, action horizon $H=10$, $S=10$ denoising steps, and $N=30$ language tokens per request at $k=5$ steps/frame.

\begin{figure}[t]
  \centering
  \begin{minipage}[c]{0.38\linewidth}
    \centering
    \includegraphics[width=0.42\linewidth]{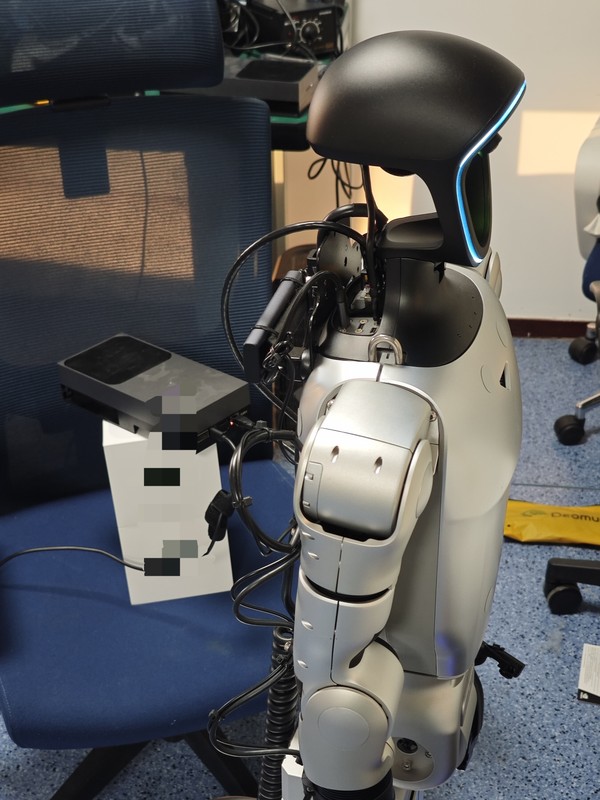}
  \end{minipage}\hfill
  \begin{minipage}[c]{0.6\linewidth}
    \centering
    \small
    \setlength{\tabcolsep}{4pt}
    \renewcommand{\arraystretch}{1.05}
    \begin{tabular}{l|rr}
      \toprule
      Stage & Baseline (ms) & \nickname (ms) \\
      \midrule
      Video stream & 9.4 & - \\
      \midrule
      Action path & 207.5 & 198.0 \\
      Language path & \textbf{822.3} & \textbf{195.4} \\
      \midrule
      Action comm. & 0.6 & - \\
      Action exec. & 333.0 & - \\
      \bottomrule
    \end{tabular}
  \end{minipage}
  \caption{Humanoid robot with on-board Jetson AGX Thor (left) and measured stage latencies (right). The action path includes one prefill and action denoising in both methods. The language path includes a second prefill and autoregressive decoding in Baseline; \nickname reuses the action-path prefill for batched decoding. The language-path reduction therefore reflects both KV sharing and batching.}
  \label{fig:realrobot}
\end{figure}

\cref{fig:realrobot} shows the platform and measured stage latencies. Summing the action and language paths gives 1030~ms for isolated execution ($207.5+822.3$) and 393~ms for \nickname ($198.0+195.4$). Once actions are dispatched, the 195~ms of language decoding can be hidden within the 333~ms of physical action execution.

\section{Conclusion}

We present \nickname, an efficient inference system for MoT VLAs under multi-task parallelism.
By proposing unified KV cache management, a novel inference design that treats the KV cache as a shared resource across tasks and over time, we enable cross-task KV sharing and cross-frame continuous batching.
For $\pi_{0.5}$, \nickname achieves up to $3.7\times$ speedup on NVIDIA GeForce RTX 4090 and Jetson AGX Thor. Additional experiments on Xiaomi-Robotics-0 and StarVLA demonstrate its applicability across backbones and multi-expert configurations.

\paragraph{Limitations.}
Our evaluation focuses on concurrent action and language generation. Further validation is needed for unified models with additional output modalities such as video; integration with asynchronous control and model-level optimizations; and end-to-end real-world applications spanning data collection, multi-task training, and closed-loop deployment.

\paragraph{Broader impacts.}
By reducing computation and energy costs, \nickname can make multi-task robotic systems more accessible. Incorrect actions or language outputs can pose risks to people interacting with robots, motivating task-specific testing and supervision in real-world deployments.

\section*{Acknowledgments and Disclosure of Funding}

We thank Bingjia Huang and Xiang Wang for their help with the humanoid robot experiments.
This work was supported in part by NSFC No. 62402280, Tsinghua University (AIR)-AsiaInfo Technologies (China), Inc. Joint Research Center for 6G Network and Intelligent Computing (20233910006), and Wuxi Research Institute of Applied Technologies, Tsinghua University under Grant 20242001120.

\bibliographystyle{oxygen-unsrtnat}
\bibliography{main}

\clearpage
\appendix

\section{Method Details}
\label{sec:method-appendix}

This section records the formal definitions and algorithm deferred from \cref{sec:method}.

\subsection{Generation Formulations for Language and Action Experts}
\label{sec:appendix-generation}

Given the prefill KV representation $\mathcal{K}_t$ from \cref{eq:kv-cache}, the language expert generates text autoregressively. For an output of $N$ tokens,
\begin{equation}
  p_{\Theta_{\text{Lang}}}(\mathbf{y}_t \mid \mathcal{K}_t) = \prod_{j=1}^{N} p_{\Theta_{\text{Lang}}}(y_{t,j} \mid y_{t,1:j-1}, \mathcal{K}_t),
  \label{eq:language-generation}
\end{equation}
while the action expert generates the action chunk $\mathbf{A}_t = \{\mathbf{a}_{t,i}\}_{i=1}^{H}$ through an iterative denoising process over $S$ steps:
\begin{equation}
  \mathbf{A}_t = \text{Denoise}_{\Theta_{\text{Act}}}^{(S)}(\boldsymbol{\epsilon}, \mathcal{K}_t), \quad \boldsymbol{\epsilon} \sim \mathcal{N}(\mathbf{0}, \mathbf{I}),
  \label{eq:action-generation}
\end{equation}
where each denoising step conditions on the prefill representation $\mathcal{K}_t$. This process can be implemented via diffusion or flow matching.

\subsection{Unified KV Cache Manager: Interface Details}
\label{sec:appendix-manager}

We expand the manager interfaces summarized in \cref{sec:unified-kv}.

\paragraph{Interface for state persistence.}
The manager $\mathcal{M}$ exposes four core operations to persist and retrieve generation states $\sigma_t$ defined in \cref{eq:state}:
\AIRinterfaces{%
  \AIRoperation{\mathcal{M}.\textsc{Store}(\sigma_t)}{r_t}{persists state $\sigma_t$ and returns request ID $r_t$}%
  \AIRoperation{\mathcal{M}.\textsc{Retrieve}(r_t)}{\sigma_t}{fetches state by ID $r_t$}%
  \AIRoperation{\mathcal{M}.\textsc{Update}(r_t, \sigma_t')}{\emptyset}{replaces existing state of ID $r_t$ with $\sigma_t'$}%
  \AIRoperation{\mathcal{M}.\textsc{Remove}(r_t)}{\emptyset}{evicts finished request state by ID $r_t$}%
}
For the one-new-request-per-frame setting in \cref{alg:per-frame}, $r_t$ identifies the request created at frame $t$.
The updated state $\sigma_t' = (\mathcal{K}_t', \mathbf{y}_t', \delta_t')$ records progress after up to $k$ decoding steps: $\mathcal{K}_t'$ extends the request cache with the new token KVs, $\mathbf{y}_t'$ appends the emitted tokens, and $\delta_t'$ is set to 1 at EOS or length $N$. The prefill portion remains unchanged.

\paragraph{Interface for batched decoding.}
At any given time, the manager maintains a set of active requests $\mathcal{R} = \{r_{t_1}, r_{t_2}, \ldots, r_{t_m}\}$, where each $r_{t_i}$ is the request ID of a request initiated at frame $t_i$ that has not yet terminated ($\delta_{t_i} = 0$). When a new request $r_t$ is created at frame $t$, it is immediately added to $\mathcal{R}$ and participates in batched decoding.
To enable efficient batched decoding across all $m = |\mathcal{R}|$ active requests, the manager defines a \emph{batched state}:
\begin{equation}
  \hat{\sigma} = \bigl(\hat{\mathcal{K}},\; \hat{\mathbf{y}},\; \hat{\boldsymbol{\delta}}\bigr),
  \label{eq:batched-state}
\end{equation}
where $\hat{\mathcal{K}} = \{(\hat{\mathbf{K}}_l, \hat{\mathbf{V}}_l)\}_{l=1}^{L}$ stacks the request caches $\{\mathcal{K}_{t_i}\}_{i=1}^{m}$ along the batch dimension at each layer $l$, $\hat{\mathbf{y}} = [\mathbf{y}_{t_1}; \mathbf{y}_{t_2}; \ldots; \mathbf{y}_{t_m}]$ concatenates token buffers, and $\hat{\boldsymbol{\delta}} = [\delta_{t_1}, \delta_{t_2}, \ldots, \delta_{t_m}]$ collects termination flags. Each request retains its own token positions and attention mask. The implementation batches states with matching prefix and cache capacities, so requests at different decoding positions can share a forward pass. The manager provides two operations to convert between individual and batched states:
\AIRinterfaces{%
  \AIRoperation{\mathcal{M}.\textsc{Batch}(\{\sigma_{t_i}\}_{i=1}^{m})}{\hat{\sigma}}{concatenates states into batched state $\hat{\sigma}$}%
  \AIRoperation{\mathcal{M}.\textsc{UnBatch}(\hat{\sigma}')}{\{\sigma_{t_i}'\}_{i=1}^{m}}{splits batched state into individual states}%
}
The batched state $\hat{\sigma}$ enables the VLM to perform autoregressive decoding on all $m$ requests in parallel. At each step, each request uses its own cache in $\hat{\mathcal{K}}$ and its current token to predict the next token. The manager separates the updated states afterward and removes completed requests.

\paragraph{Request storage and allocation.}
\label{sec:appendix-memory}

Each language request keeps a private state containing its prefill context, generated tokens, and termination flag. The request cache reserves capacity for up to $N$ decoded tokens when the request starts and keeps that capacity until completion. The action expert reads the prefill representation independently during denoising, while the language state is retained across frames and updated after each batched decode. Removing a completed request releases its state so the framework allocator can reuse the storage.

\subsection{Per-Frame Execution Algorithm}
\label{sec:appendix-algorithm}

\cref{alg:per-frame} formalizes the per-frame execution flow described in \cref{sec:unified-kv}. At each frame, the system processes one new observation (spawning a language generation request and producing actions) while continuing $m-1$ in-flight language requests from previous frames, resulting in $m$ total active requests processed in parallel.

\begin{algorithm}[t]
  \caption{Per-Frame Execution Flow}
  \label{alg:per-frame}
  \KwIn{Frame index $t$, new observation $\mathbf{o}_t$, active request IDs $\mathcal{R} = \{r_{t_i}\}_{i=1}^{m-1}$, manager $\mathcal{M}$, per-frame decoding budget $k$, action denoising steps $S$}
  \KwOut{Actions $\mathbf{A}_t$, updated language tokens $\{\mathbf{y}_{t_i}\}_{i=1}^{m}$, active request IDs $\mathcal{R}$.}
  \BlankLine
  \tcp{New request: generate actions and initialize language}
  $\mathcal{K}_t \leftarrow \textsc{Prefill}(\mathbf{o}_t)$\;
  $\mathbf{A}_t \leftarrow \textsc{ActionDenoise}(\mathcal{K}_t, S)$\;
  $\sigma_t \leftarrow \textsc{InitState}(\mathcal{K}_t)$\;
  $r_t \leftarrow \mathcal{M}.\textsc{Store}(\sigma_t)$\;
  \BlankLine
  \tcp{Batched decode: advance each active request by up to $k$ tokens}
  $\mathcal{R} \leftarrow \mathcal{R} \cup \{r_t\}$ \tcp*{Add new request to active set, now $|\mathcal{R}| = m$}
  $\{\sigma_{t_i}\}_{i=1}^{m} \leftarrow \{\mathcal{M}.\textsc{Retrieve}(r_{t_i}) \mid r_{t_i} \in \mathcal{R}\}$\;
  $\hat{\sigma} \leftarrow \mathcal{M}.\textsc{Batch}(\{\sigma_{t_i}\}_{i=1}^{m})$\;
  $\hat{\sigma}' \leftarrow \textsc{BatchedLanguageDecode}(\hat{\sigma}, k)$\;
  $\{\sigma_{t_i}'\}_{i=1}^{m} \leftarrow \mathcal{M}.\textsc{UnBatch}(\hat{\sigma}')$\;
  \BlankLine
  \tcp{Extract language outputs and update or evict requests}
  \ForEach{$i = 1$ \KwTo $m$}{
    $\mathbf{y}_{t_i} \leftarrow \sigma_{t_i}'.y$ \tcp*{Get updated tokens per request}
    \If{$\sigma_{t_i}'.\delta = 1$}{
      $\mathcal{M}.\textsc{Remove}(r_{t_i})$\;
      $\mathcal{R} \leftarrow \mathcal{R} \setminus \{r_{t_i}\}$\;
    }
    \lElse{$\mathcal{M}.\textsc{Update}(r_{t_i}, \sigma_{t_i}')$}
  }
\end{algorithm}

\section{Evaluation of \texorpdfstring{$\pi_{0.5}$}{pi0.5}}
\label{sec:appendix-experiments}

This section supplements the $\pi_{0.5}$ evaluation in \cref{sec:experiments} with cross-configuration speed results, action quality, and resource use.

\subsection{End-to-End Performance}
\label{sec:appendix-e2e}

We extend the main LIBERO results to ALOHA and DROID, then examine how output length $N$ and decoding budget $k$ affect performance.

\paragraph{End-to-End Tradeoff on ALOHA and DROID.}
\label{sec:appendix-e2e-tradeoff}

\begin{figure}[t]
  \centering
  \begin{subfigure}[t]{\linewidth}
    \centering
    \includegraphics[width=\linewidth]{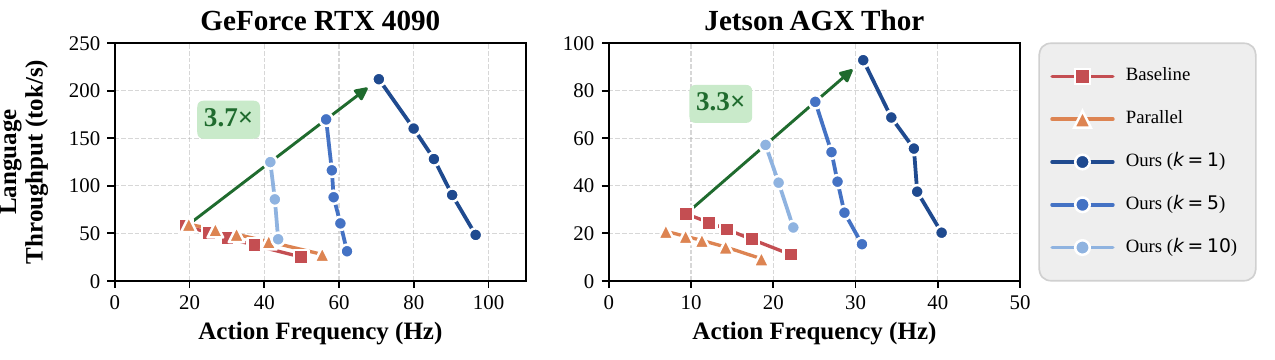}
    \caption{ALOHA.}
    \label{fig:e2e-tradeoff-aloha}
  \end{subfigure}
  \begin{subfigure}[t]{\linewidth}
    \centering
    \includegraphics[width=\linewidth]{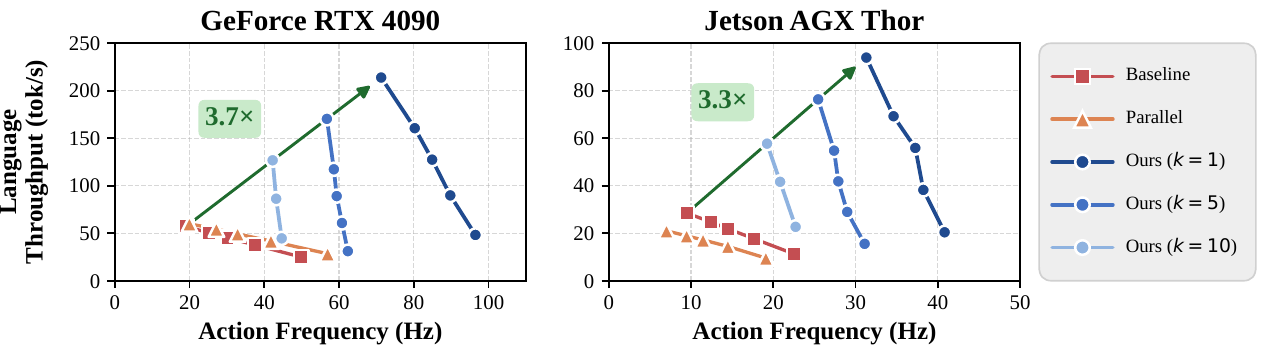}
    \caption{DROID.}
    \label{fig:e2e-tradeoff-droid}
  \end{subfigure}
  \caption{Action-frequency vs.\ language-throughput tradeoff on the ALOHA and DROID configurations, sweeping total decoding steps $N$ and decoding steps per frame $k$. Trends mirror the LIBERO results in \cref{fig:e2e-tradeoff}: \nickname extends the Pareto frontier beyond what either baseline can reach.}
  \label{fig:e2e-tradeoff-aloha-droid}
\end{figure}

\cref{fig:e2e-tradeoff-aloha-droid} extends the main tradeoff to ALOHA and DROID. In both configurations, \nickname moves the Pareto frontier outward, matching the LIBERO trend across configurations with different observation and action spaces.

\paragraph{Effect of Language Output Length.}
\label{sec:appendix-e2e-heatmap}

\begin{figure}[t]
  \centering
  \includegraphics[width=\linewidth]{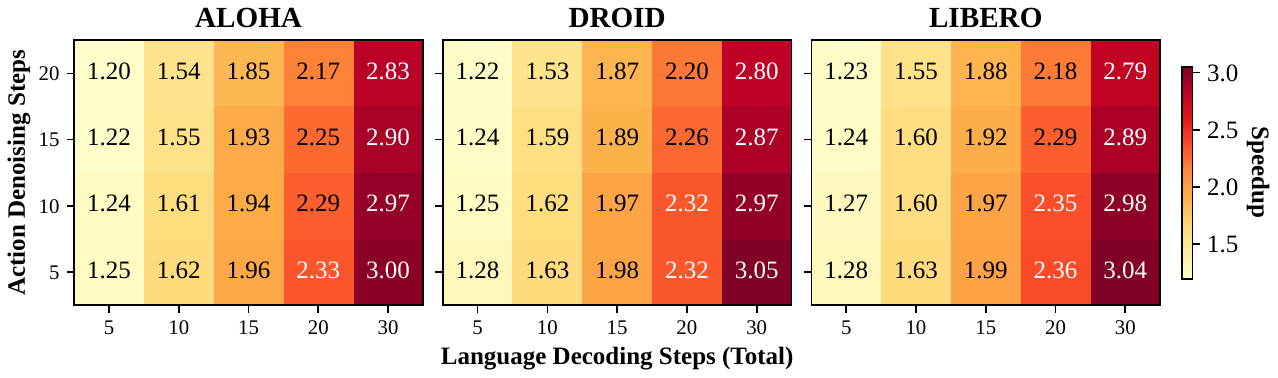}
  \caption{Speedup ratio for action frequency of \nickname vs.\ baseline across LIBERO, ALOHA, and DROID configurations (subplots), action denoising steps $S$ (y-axis), and total decoding steps $N$ (x-axis), with decoding steps per frame fixed at $k=5$. The speedup increases with larger $N$, while the configuration choice and $S$ have smaller impact.}
  \label{fig:e2e-speedup-heatmap-decoding-appendix}
\end{figure}

\cref{fig:e2e-speedup-heatmap-decoding-appendix} fixes $k=5$ and varies total decoding steps $N$, complementing the main heatmap that fixes $N=30$. The speedup grows with $N$ across all three configurations, as longer requests create more opportunities for cross-frame batching.

\paragraph{Output Length and Decoding Budget.}
\label{sec:appendix-e2e-latency}

\begin{figure}[t]
  \centering
  \begin{subfigure}[t]{\linewidth}
    \centering
    \includegraphics[width=\linewidth]{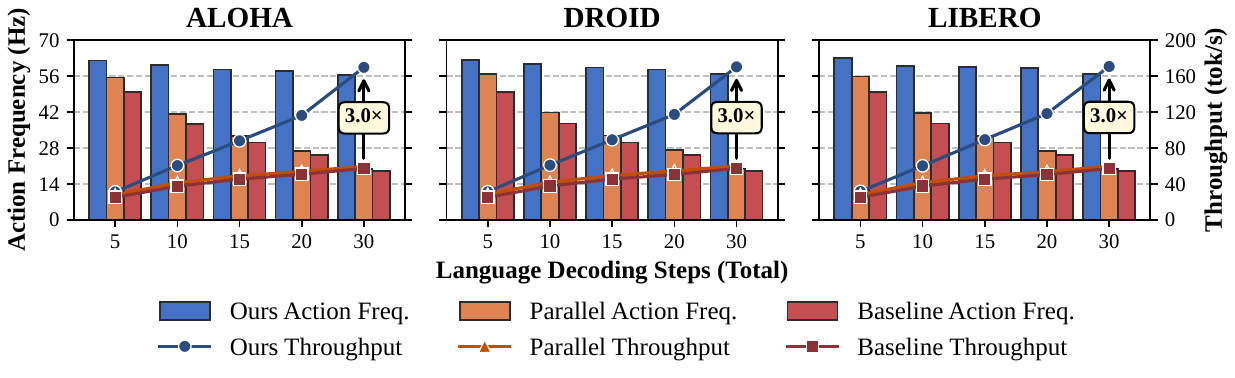}
    \caption{Varying total decoding steps (steps per frame $k=5$).}
    \label{fig:e2e-latency-throughput-vs-decoding-steps-appendix}
  \end{subfigure}
  \begin{subfigure}[t]{\linewidth}
    \centering
    \includegraphics[width=\linewidth]{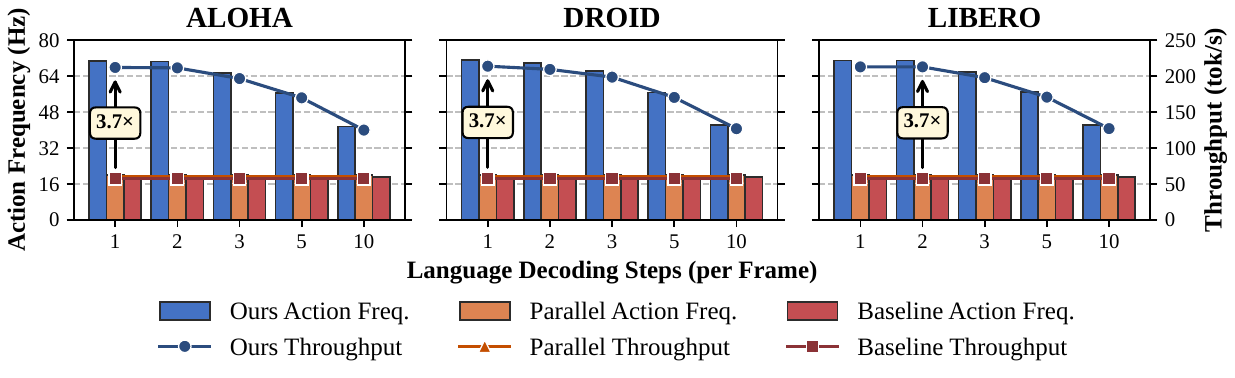}
    \caption{Varying steps per frame (total decoding steps $N=30$).}
    \label{fig:e2e-latency-throughput-vs-steps-per-frame-appendix}
  \end{subfigure}
  \caption{Action-frequency and language-throughput curves. Panel (a) varies total decoding steps $N$ at fixed $k=5$, and panel (b) varies decoding steps per frame $k$ at fixed $N=30$. \nickname reaches over 200~tokens/s and 70~Hz simultaneously.}
  \label{fig:e2e-latency-throughput-appendix}
\end{figure}

\cref{fig:e2e-latency-throughput-appendix} plots action frequency and language throughput separately as $N$ or $k$ varies, showing how each contributes to the tradeoff.

\subsection{Action Quality}
\label{sec:appendix-quality}

\cref{tab:quality} reports evaluation over ten environment and policy-sampling seeds. Each seed covers four suites, ten tasks per suite, and fifty rollouts per task, for 20,000 rollouts in total.
We compute the mean and sample standard deviation over the ten seed-level success rates. The 95\% confidence interval is $\bar{s}\pm t_{0.975,9}\,s/\sqrt{10}$, using Student's $t$ distribution with nine degrees of freedom. For the overall result, each seed contributes its success rate over all 2,000 rollouts. The openpi column contains published point estimates.

\begin{table}[t]
  \centering
  \small
  \setlength{\tabcolsep}{4pt}
  \renewcommand{\arraystretch}{1.05}
  \caption{LIBERO success rates (\%). \nickname results are means and 95\% confidence intervals across ten seeds; openpi values are published point estimates.}
  \label{tab:quality}
  \begin{tabular}{lrrr}
    \toprule
    Suite & openpi & \nickname mean & \nickname 95\% CI \\
    \midrule
    Spatial & 98.80 & 98.42 & [98.12, 98.72] \\
    Object & 98.20 & 98.52 & [97.95, 99.09] \\
    Goal & 98.00 & 97.24 & [96.66, 97.82] \\
    LIBERO-10 & 92.40 & 92.86 & [92.40, 93.32] \\
    Overall & 96.85 & 96.76 & [96.55, 96.97] \\
    \bottomrule
  \end{tabular}
\end{table}

\subsection{Memory and Energy}
\label{sec:appendix-overhead}

\cref{tab:overhead} reports the memory and energy measurements supporting the headline numbers in \cref{sec:exp-quality-overhead}. Energy is computed as average device power times mean frame duration in seconds. At steady state with one new request per frame, this is also the energy per completed request.

\begin{table}[t]
  \centering
  \small
  \setlength{\tabcolsep}{4pt}
  \renewcommand{\arraystretch}{1.05}
  \caption{Comparison of memory and energy cost. \nickname adds modest memory overhead while achieving substantial energy savings.}
  \label{tab:overhead}
  \begin{tabular}{l|rrrr}
    \toprule
    \shortstack[c]{Setting\\~} & \shortstack{Peak\\Mem. (GiB)} & \shortstack{Avg.\\Power (W)} & \shortstack{Energy\\/Req. (J)} & \shortstack{$\Delta$ Energy\\ /Req. (\%)} \\
    \midrule
    Baseline & 6.43 & 293.5 & 117.4 & 0 \\
    Parallel & 12.49 & 324.4 & 120.9 & $+$3.0 \\
    Ours w/o Batching & 6.43 & 287.9 & 97.7 & \textbf{$-$16.7} \\
    Ours (batch: 2) & 7.35 & 173.3 & 39.1 & \textbf{$-$66.7} \\
    Ours (batch: 4) & 7.41 & 154.4 & 25.8 & \textbf{$-$78.0} \\
    \bottomrule
  \end{tabular}
\end{table}

\section{Evaluation across Models and Configurations}
\label{sec:appendix-models}

These experiments extend \cref{sec:exp-models} by examining how the gains from prefix reuse and continuous batching vary with language output length, expert count, and backbone size. We evaluate Xiaomi-Robotics-0 and StarVLA's Qwen3VL-PI\_v3 on RTX 4090, using $S=5$ and $S=4$ denoising steps, respectively. We then add GR00T and OFT action experts through StarVLA and evaluate Qwen3.5 backbones of different sizes.

Following \cref{sec:ablation}, we compare Baseline, Ours w/o Batching, and Ours under the same inputs and generation settings within each configuration. One language request arrives per frame and generates $N$ tokens. We sweep $N\in\{5,10,15,20,30\}$ and $k\in\{1,5,10\}$ where $N$ is divisible by $k$. \cref{tab:cross-model-results} shows results for $N\in\{10,20,30\}$.

\paragraph{Backbones and expert count.}

Across Xiaomi-Robotics-0 and the StarVLA configurations, longer language outputs and smaller per-frame decoding budgets increase the benefit of continuous batching (\cref{tab:cross-model-results}). With one new request per frame, increasing $N/k$ creates larger batches that amortize language decoding across requests. \nickname reduces frame latency in every tested setting with $N>k$. When $N=k$, each request finishes within one frame, leaving no requests to batch across frames. At $N=k=10$ on Xiaomi-Robotics-0, \nickname is slightly slower than Baseline (0.96$\times$), while Ours w/o Batching provides a 1.16$\times$ speedup.

Adding action experts increases the benefit of prefix reuse. In StarVLA, Ours w/o Batching saves approximately 25, 50, and 75~ms per frame with two, three, and four experts, respectively (\cref{tab:cross-model-results}). Each additional expert would otherwise repeat the same prefill. These savings remain nearly constant across language output lengths, while the benefit of continuous batching grows with $N/k$.

\begin{table}[!htbp]
  \centering
  \small
  \setlength{\tabcolsep}{4pt}
  \renewcommand{\arraystretch}{1.05}
  \caption{Cross-model and multi-expert results on RTX 4090. Baseline reports frame latency in ms; all other method columns report speedup over Baseline. $E$ counts language and action experts. For StarVLA, $E=2$ is Qwen3VL-PI\_v3, with GR00T added at $E=3$ and OFT at $E=4$.}
  \label{tab:cross-model-results}
  \begin{tabular}{lrrrrrrr}
    \toprule
    Model & $E$ & $N$ & \shortstack[r]{Baseline\\(ms)} & \shortstack[r]{Ours w/o\\Batching} & \shortstack[r]{Ours\\($k=1$)} & \shortstack[r]{Ours\\($k=5$)} & \shortstack[r]{Ours\\($k=10$)} \\
    \midrule
    \multirow{3}{*}{Xiaomi-Robotics-0} & \multirow{3}{*}{2} & 10 & 249.9 & 1.16$\times$ & \textbf{2.91$\boldsymbol{\times}$} & 1.49$\times$ & 0.96$\times$ \\
     &  & 20 & 406.3 & 1.09$\times$ & \textbf{4.47$\boldsymbol{\times}$} & 2.39$\times$ & 1.47$\times$ \\
     &  & 30 & 562.8 & 1.07$\times$ & \textbf{5.83$\boldsymbol{\times}$} & 3.23$\times$ & 2.03$\times$ \\
    \midrule
    \multirow{3}{*}{\shortstack[l]{StarVLA Qwen3VL-PI\_v3}} & \multirow{3}{*}{2} & 10 & 211.8 & 1.13$\times$ & \textbf{3.20$\boldsymbol{\times}$} & 1.71$\times$ & 1.13$\times$ \\
     &  & 20 & 344.9 & 1.08$\times$ & \textbf{5.00$\boldsymbol{\times}$} & 2.73$\times$ & 1.73$\times$ \\
     &  & 30 & 478.0 & 1.05$\times$ & \textbf{6.64$\boldsymbol{\times}$} & 3.60$\times$ & 2.35$\times$ \\
    \midrule
    \multirow{3}{*}{\shortstack[l]{StarVLA Qwen3VL-PI\_v3\\(+ GR00T)}} & \multirow{3}{*}{3} & 10 & 250.7 & 1.24$\times$ & \textbf{3.13$\boldsymbol{\times}$} & 1.82$\times$ & 1.25$\times$ \\
     &  & 20 & 383.6 & 1.15$\times$ & \textbf{4.64$\boldsymbol{\times}$} & 2.74$\times$ & 1.80$\times$ \\
     &  & 30 & 515.8 & 1.11$\times$ & \textbf{6.05$\boldsymbol{\times}$} & 3.53$\times$ & 2.38$\times$ \\
    \midrule
    \multirow{3}{*}{\shortstack[l]{StarVLA Qwen3VL-PI\_v3\\(+ GR00T + OFT)}} & \multirow{3}{*}{4} & 10 & 276.3 & 1.37$\times$ & \textbf{3.44$\boldsymbol{\times}$} & 2.01$\times$ & 1.37$\times$ \\
     &  & 20 & 409.1 & 1.22$\times$ & \textbf{4.91$\boldsymbol{\times}$} & 2.91$\times$ & 1.92$\times$ \\
     &  & 30 & 541.2 & 1.16$\times$ & \textbf{6.30$\boldsymbol{\times}$} & 3.69$\times$ & 2.49$\times$ \\
    \bottomrule
  \end{tabular}
\end{table}

\paragraph{Model size.}
\label{sec:appendix-scaling}

We evaluate Qwen3.5 backbones from 0.8B to 9B through StarVLA on RTX 4090. Each backbone uses a matching PI\_v3 action head that generates 16 actions with $S=4$ denoising steps. The observation and task text are fixed across sizes. \nickname provides similar speedups across model sizes (\cref{tab:model-scaling}). At $N=30,k=5$, speedup ranges from 3.58$\times$ to 3.79$\times$ as the backbone grows from 0.8B to 9B, while Baseline latency rises from 429.0 to 731.9~ms. All four sizes also follow the workload trend described above: longer outputs and smaller per-frame budgets increase the benefit of batching.

Batched decoding remains efficient for the larger backbones. Increasing the batch size from 1 to 30 raises one-token decode latency from 16.0 to 18.2~ms for 4B and from 20.7 to 27.2~ms for 9B. Processing more requests together therefore adds relatively little time to each decoding step.

\begin{table}[!htbp]
  \centering
  \small
  \setlength{\tabcolsep}{4pt}
  \renewcommand{\arraystretch}{1.05}
  \caption{Qwen3.5 model-size comparison on RTX 4090. Baseline reports frame latency in ms; all other columns report speedup over Baseline.}
  \label{tab:model-scaling}
  \begin{tabular}{lrrrrrr}
    \toprule
    Backbone & $N$ & \shortstack[r]{Baseline\\(ms)} & \shortstack[r]{Ours w/o\\Batching} & \shortstack[r]{Ours\\($k=1$)} & \shortstack[r]{Ours\\($k=5$)} & \shortstack[r]{Ours\\($k=10$)} \\
    \midrule
    \multirow{3}{*}{0.8B} & 10 & 196.4 & 1.15$\times$ & \textbf{2.91$\boldsymbol{\times}$} & 1.65$\times$ & 1.12$\times$ \\
     & 20 & 313.6 & 1.10$\times$ & \textbf{4.62$\boldsymbol{\times}$} & 2.62$\times$ & 1.70$\times$ \\
     & 30 & 429.0 & 1.06$\times$ & \textbf{6.30$\boldsymbol{\times}$} & 3.58$\times$ & 2.32$\times$ \\
    \midrule
    \multirow{3}{*}{2B} & 10 & 211.3 & 1.16$\times$ & \textbf{2.99$\boldsymbol{\times}$} & 1.71$\times$ & 1.14$\times$ \\
     & 20 & 331.2 & 1.09$\times$ & \textbf{4.65$\boldsymbol{\times}$} & 2.65$\times$ & 1.73$\times$ \\
     & 30 & 453.3 & 1.07$\times$ & \textbf{6.31$\boldsymbol{\times}$} & 3.62$\times$ & 2.34$\times$ \\
    \midrule
    \multirow{3}{*}{4B} & 10 & 274.0 & 1.15$\times$ & \textbf{3.03$\boldsymbol{\times}$} & 1.70$\times$ & 1.12$\times$ \\
     & 20 & 437.9 & 1.09$\times$ & \textbf{4.82$\boldsymbol{\times}$} & 2.72$\times$ & 1.76$\times$ \\
     & 30 & 602.9 & 1.06$\times$ & \textbf{6.55$\boldsymbol{\times}$} & 3.71$\times$ & 2.39$\times$ \\
    \midrule
    \multirow{3}{*}{9B} & 10 & 316.7 & 1.13$\times$ & \textbf{3.22$\boldsymbol{\times}$} & 1.68$\times$ & 1.13$\times$ \\
     & 20 & 523.6 & 1.07$\times$ & \textbf{5.23$\boldsymbol{\times}$} & 2.73$\times$ & 1.72$\times$ \\
     & 30 & 731.9 & 1.05$\times$ & \textbf{7.18$\boldsymbol{\times}$} & 3.79$\times$ & 2.39$\times$ \\
    \bottomrule
  \end{tabular}
\end{table}

\paragraph{Tensor parallelism.}
We also evaluate the 9B model using one or two RTX 4090 GPUs on a separate eight-GPU server, connected through PCIe without NVLink. At $N=30,k=5$, \nickname provides 3.50$\times$ speedup at TP=1 and 3.71$\times$ at TP=2. Its frame latency increases from 493.4 to 535.3~ms with the second GPU. Thus, \nickname retains its benefit under tensor parallelism, although using two GPUs does not reduce latency in this setup.

\section{Runtime Analysis}

The systems experiments in this section use $\pi_{0.5}$-LIBERO on RTX 4090.

\subsection{Prefix-Reuse Benefits}
\label{sec:appendix-small-n}

Prefix computation accounts for a substantial part of inference when language outputs are short. With $S=10$, one prefill takes about 44.0~ms, more than twice the 20.4~ms spent on action denoising (\cref{tab:small-n-profile}). Prefill processes the image and task instruction through the VLM backbone, whereas denoising runs the smaller action expert. Reusing this prefill for language generation removes the second prefix computation in Baseline.

Ours w/o Batching reduces end-to-end frame latency by 36.4\% at $N=1$ and 25.9\% at $N=5$. As language outputs grow, the absolute saving stays near 47--48~ms, while its relative contribution falls to 9.4\% at $N=30$. Prefix reuse therefore provides its largest relative benefit for short outputs; continuous batching addresses the growing decoding cost for longer outputs.

\begin{table}[!htbp]
  \centering
  \small
  \setlength{\tabcolsep}{4pt}
  \renewcommand{\arraystretch}{1.05}
  \caption{Inference latency breakdown and prefix-reuse benefits for $\pi_{0.5}$-LIBERO ($S=10$). The breakdown reports one prefill, action denoising, and $N$ language decoding steps in Ours w/o Batching. The final column gives its end-to-end latency reduction relative to Baseline.}
  \label{tab:small-n-profile}
  \begin{tabular}{rrrrr}
    \toprule
    $N$ & Prefill (ms) & Denoising (ms) & Decoding (ms) & E2E reduction (\%) \\
    \midrule
    1 & 44.04 & 20.50 & 14.51 & \textbf{36.4} \\
    5 & 44.04 & 20.46 & 66.61 & \textbf{25.9} \\
    10 & 43.99 & 20.39 & 132.03 & \textbf{19.2} \\
    20 & 44.01 & 20.41 & 263.15 & \textbf{12.5} \\
    30 & 44.22 & 20.49 & 393.85 & \textbf{9.4} \\
    \bottomrule
  \end{tabular}
\end{table}

\subsection{KV-Management Overhead}
\label{sec:appendix-manager-overhead}

We measure the cost of maintaining the language batch in JAX and PyTorch at $S=10$ and $N=30$, with one new request per frame. Setting $k=5$ or $k=1$ gives active batches of 6 or 30 requests. Management includes stacking request caches, splitting the updated batch, and request bookkeeping. We report management time as a fraction of frame latency (\cref{tab:manager-overhead}).

Cache splitting accounts for most of the management cost in both implementations. At batch size 6, management takes 3.4\% of the frame in JAX and 1.2\% in PyTorch. At batch size 30, these shares rise to 17.5\% and 6.6\%, respectively, as more request caches must be assembled and separated. Bookkeeping remains below 0.02~ms. Persistent cache slots could reduce the cost of stacking and splitting.

The larger batch still reduces overall frame latency because $k=1$ requires fewer language decoding steps per frame. Frame latency falls from 147.6 to 120.6~ms in JAX and from 168.3 to 148.1~ms in PyTorch.

\begin{table}[!htbp]
  \centering
  \small
  \setlength{\tabcolsep}{4pt}
  \renewcommand{\arraystretch}{1.05}
  \caption{KV-management overhead for $\pi_{0.5}$-LIBERO on RTX 4090 ($S=10$, $N=30$). Manager total includes stacking, splitting, and bookkeeping. Manager share is the percentage of frame latency spent on these operations.}
  \label{tab:manager-overhead}
  \begin{tabular}{lrrrr}
    \toprule
    & \multicolumn{2}{c}{JAX} & \multicolumn{2}{c}{PyTorch} \\
    \cmidrule(lr){2-3}\cmidrule(lr){4-5}
    Measurement & \shortstack[r]{Batch 6\\($k=5$)} & \shortstack[r]{Batch 30\\($k=1$)} & \shortstack[r]{Batch 6\\($k=5$)} & \shortstack[r]{Batch 30\\($k=1$)} \\
    \midrule
    KV stacking (ms) & 0.77 & 2.82 & 0.34 & 1.28 \\
    KV splitting (ms) & 4.24 & 18.24 & 1.73 & 8.53 \\
    Manager total (ms) & 5.01 & 21.08 & 2.08 & 9.82 \\
    Frame latency (ms) & 147.59 & 120.62 & 168.32 & 148.07 \\
    \midrule
    Manager share (\%) & \textbf{3.40} & \textbf{17.48} & \textbf{1.24} & \textbf{6.63} \\
    \bottomrule
  \end{tabular}
\end{table}

\subsection{Runtime Stability}
\label{sec:appendix-stability}

We test whether repeated request arrivals and completions cause memory use to grow over time. We run $\pi_{0.5}$-LIBERO for 3,000 measured frames on RTX 4090 under a fluctuating workload with $k=4$. Arrival counts repeat every 17 frames, while request lengths follow a seven-frame pattern. The combined workload therefore repeats every 119 frames, giving 25 complete cycles with idle periods, bursts, and up to 40 active requests.

\begin{figure}[!htbp]
  \centering
  \includegraphics[width=\linewidth]{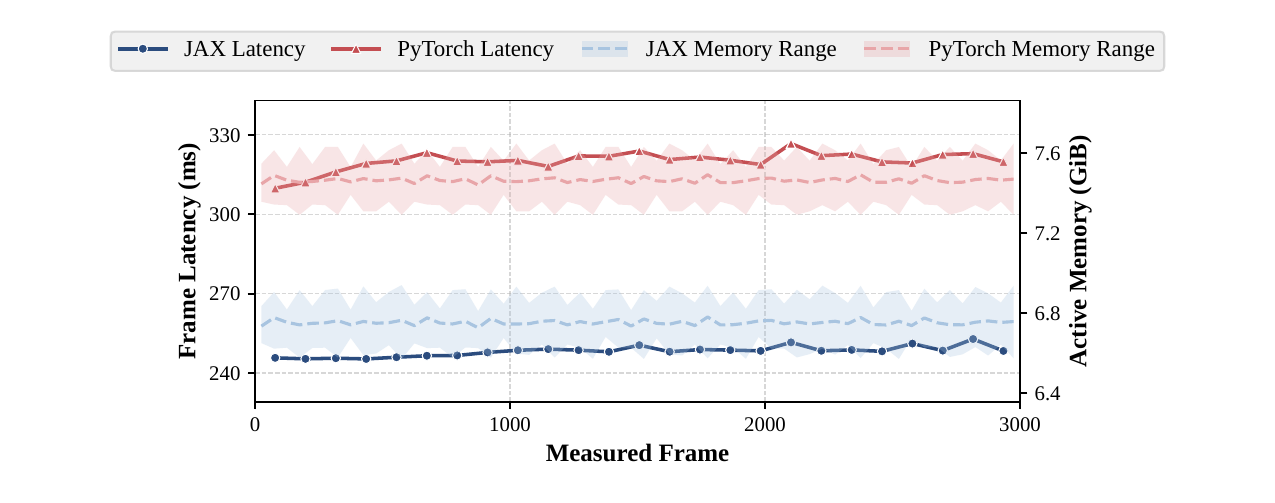}
  \caption{Memory and latency under the fluctuating $\pi_{0.5}$-LIBERO workload on RTX 4090 ($k=4$). Solid curves show the mean frame latency for each complete 119-frame workload cycle. Dashed curves and shaded bands show mean and min--max active memory in 50-frame windows.}
  \label{fig:longrun-stability}
\end{figure}

Memory use remains bounded throughout the run (\cref{fig:longrun-stability}). Active memory ranges from 6.57 to 6.94~GiB in JAX and from 7.29 to 7.65~GiB in PyTorch, following the changing request mix without a progressive increase. The cycle-averaged frame latency changes from 245.6 to 249.8~ms in JAX and from 315.5 to 320.9~ms in PyTorch between the first and last five complete cycles. These results show stable memory use and modest average-latency drift under this fluctuating workload; the run also records GPU temperatures of 87--90$^\circ$C, which can affect sustained latency.

\subsection{Language Completion Time}
\label{sec:appendix-budget}

Continuous batching spreads language generation across action updates. A request of length $N$ takes $\lceil N/k\rceil$ frames to complete, giving an estimated completion time of $\lceil N/k\rceil T$ at steady-state frame latency $T$. Thus, reducing $k$ can improve action frequency while increasing the time needed to complete an individual language request.

For $\pi_{0.5}$-LIBERO on RTX 4090 with $S=10$, $N=30$, $k=5$, and one new request per frame, mean frame latency is 175.8~ms, the same operating point reported in the main end-to-end comparison (\cref{sec:exp-e2e,fig:e2e-tradeoff}). A language request spans six frames, giving an estimated completion time of 1.055~s. Summing the measured frame durations for each complete request gives a median of 1.057~s, close to this estimate. This illustrates how $k$ controls the balance between action updates and language response time.

\section{Textual-Memory Application}
\label{sec:appendix-memory-demo}

We evaluate \nickname in a LIBERO application that generates textual descriptions of completed task steps alongside actions. We train a rank-16 LoRA~\cite{hu2022lora} on observation--text pairs derived from simulator predicates, using separate episodes for training and validation. The adapter applies only to language decoding; shared prefill and action generation use the frozen $\pi_{0.5}$-LIBERO model.

\begin{figure}[!htbp]
  \centering
  \includegraphics[width=\linewidth]{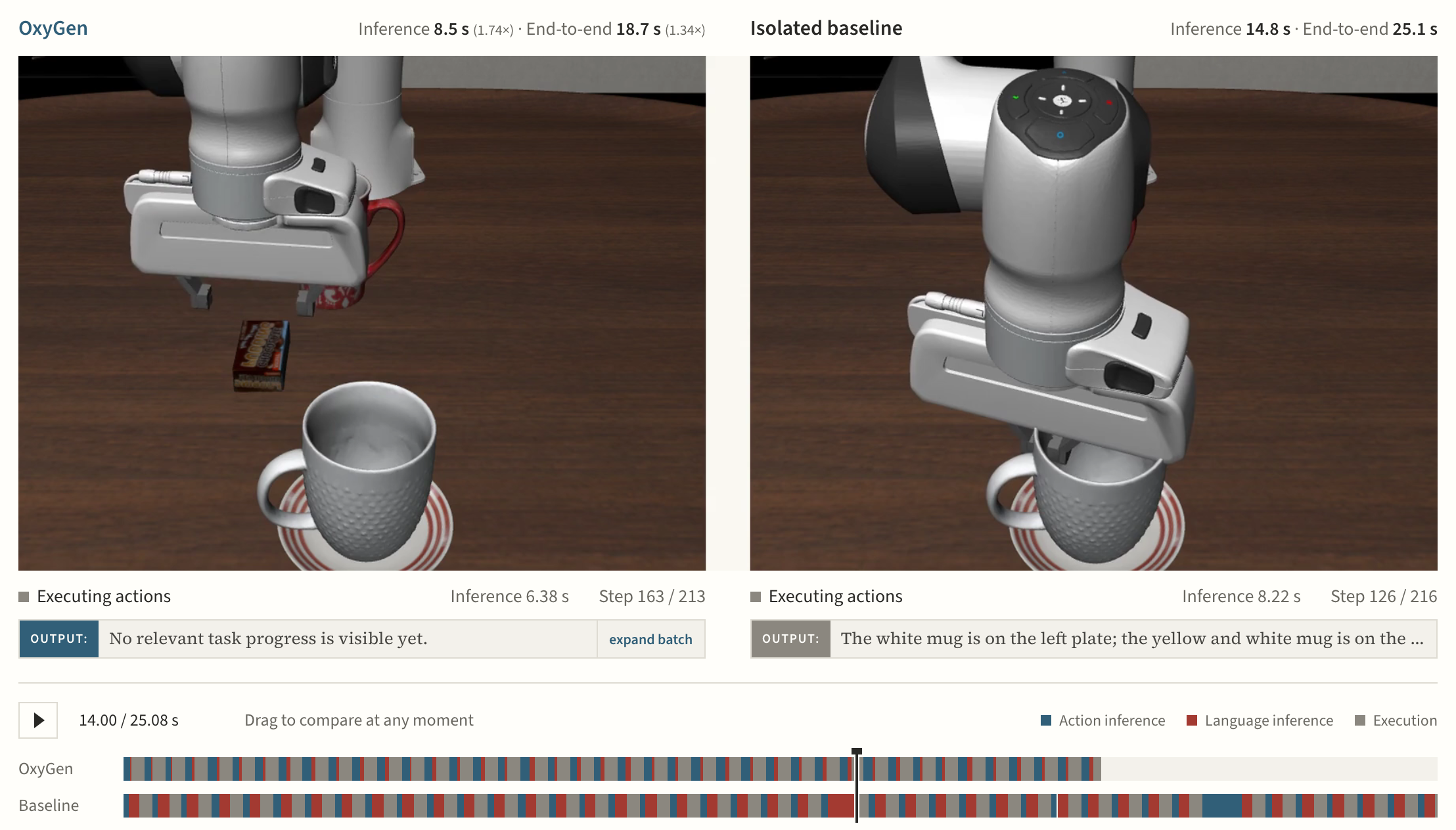}
  \caption{Textual-memory generation alongside LIBERO control. \nickname (left) makes faster task progress than isolated execution (right) while generating textual memory. The replay combines measured inference time with real-time simulator execution.}
  \label{fig:textual-memory-demo}
\end{figure}

We compare \nickname with isolated execution on the same LIBERO task (\cref{fig:textual-memory-demo}). By batching language requests across action updates, \nickname reduces the inference overhead of textual-memory generation and enables faster task progress. This example illustrates how continuous batching supports an auxiliary language task alongside ongoing control.

The project page\footnote{\url{https://air-embodied-brain.github.io/OxyGen/}} provides replays for three LIBERO tasks. We release the adapter\footnote{\url{https://huggingface.co/xxxxyu/oxygen-pi05-textual-memory-lora}} and textual annotations\footnote{\url{https://huggingface.co/datasets/xxxxyu/libero-textual-memory-annotations}}, with training and serving code in the project repository.

\end{document}